\documentclass[10pt,twocolumn,letterpaper]{article}

\usepackage{iccv}
\usepackage{times}
\usepackage{epsfig}
\usepackage{graphicx}
\usepackage{amsmath}
\usepackage{amssymb}
\usepackage{amsbsy}
\usepackage[style=base]{caption}
\usepackage{dsfont}
\usepackage{subcaption}
\usepackage{cite}
\usepackage[nomain,acronym]{glossaries}
\usepackage{verbatim}
\usepackage{siunitx}
\usepackage{xcolor}
\usepackage[accsupp]{axessibility}
\usepackage{pdfpages}

\makeatletter
\@namedef{ver@everyshi.sty}{}
\makeatother
\usepackage{tikz}

\newcommand{\vm}[1]{\protect\ensuremath\boldsymbol{#1}}

\newacronym{gcm}{GCM}{Global Context Module}
\newacronym{arm}{ARM}{Attention Refinement Module}
\newacronym{ffm}{FFM}{Feature Fusion Module}
\newacronym{dgc}{DGC}{Dense Geometrical Constrains Module}

\newcommand\copyrighttext{\footnotesize \textcopyright~2021 IEEE. Personal use of this material is permitted. Permission from IEEE must be obtained for all other uses, in any current or future media, including reprinting/republishing this material for advertising or promotional purposes, creating new collective works, for resale or redistribution to servers or lists, or reuse of any copyrighted component of this work in other works.
DOI: \href{https://doi.org/10.1109/ICCV48922.2021.01551}{10.1109/ICCV48922.2021.01551}
}

\newcommand\copyrightnotice{%
    \begin{tikzpicture}[remember picture,overlay]%
     \node[anchor=south, xshift=-8pt, yshift=20pt] at (current page.south)%
     {\fbox{\parbox{\dimexpr\textwidth-\fboxsep-\fboxrule\relax}{\copyrighttext}}};%
     \end{tikzpicture}%
}

\iccvfinalcopy

\def\iccvPaperID{8933}

\ificcvfinal\fi

\begin{document}

\title{MGNet: Monocular Geometric Scene Understanding for Autonomous Driving}

\author{
Markus Sch\"on\qquad Michael Buchholz\qquad Klaus Dietmayer \\
Institute of Measurement, Control and Microtechnology, Ulm University \\
{\tt\small \{markus.schoen,michael.buchholz,klaus.dietmayer\}@uni-ulm.de}
}

\maketitle
\ificcvfinal\thispagestyle{empty}\fi

\begin{abstract}

We introduce MGNet, a multi-task framework for monocular geometric scene understanding.
We define monocular geometric scene understanding as the combination of two known tasks: Panoptic segmentation and self-supervised monocular depth estimation.
Panoptic segmentation captures the full scene not only semantically, but also on an instance basis.
Self-supervised monocular depth estimation uses geometric constraints derived from the camera measurement model in order to measure depth from monocular video sequences only.
To the best of our knowledge, we are the first to propose the combination of these two tasks in one single model.
Our model is designed with focus on low latency to provide fast inference in real-time on a single consumer-grade GPU.
During deployment, our model produces dense 3D point clouds with instance aware semantic labels from single high-resolution camera images.
We evaluate our model on two popular autonomous driving benchmarks, i.e., Cityscapes and KITTI, and show competitive performance among other real-time capable methods.
Source code is available at \url{https://github.com/markusschoen/MGNet}.

\end{abstract}

\section{Introduction}

\begin{figure}[t]
\begin{subfigure}[b]{0.48\linewidth}
  \centering
  \includegraphics[width=\linewidth]{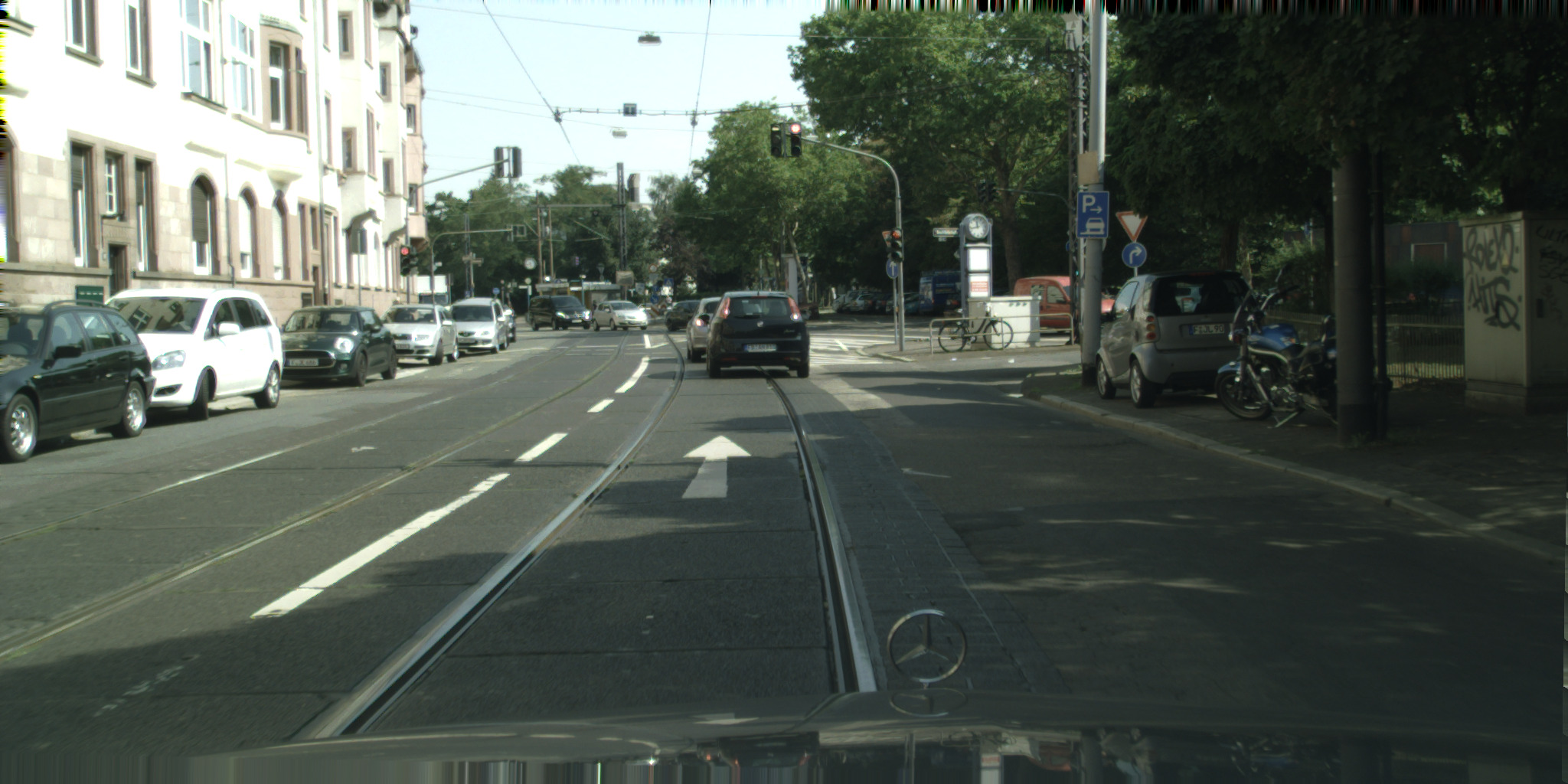}
  \caption{Input image}
  \label{fig:prediction:input}
\end{subfigure}
\hspace*{\fill}
\begin{subfigure}[b]{0.48\linewidth}
  \centering
  \includegraphics[width=\linewidth]{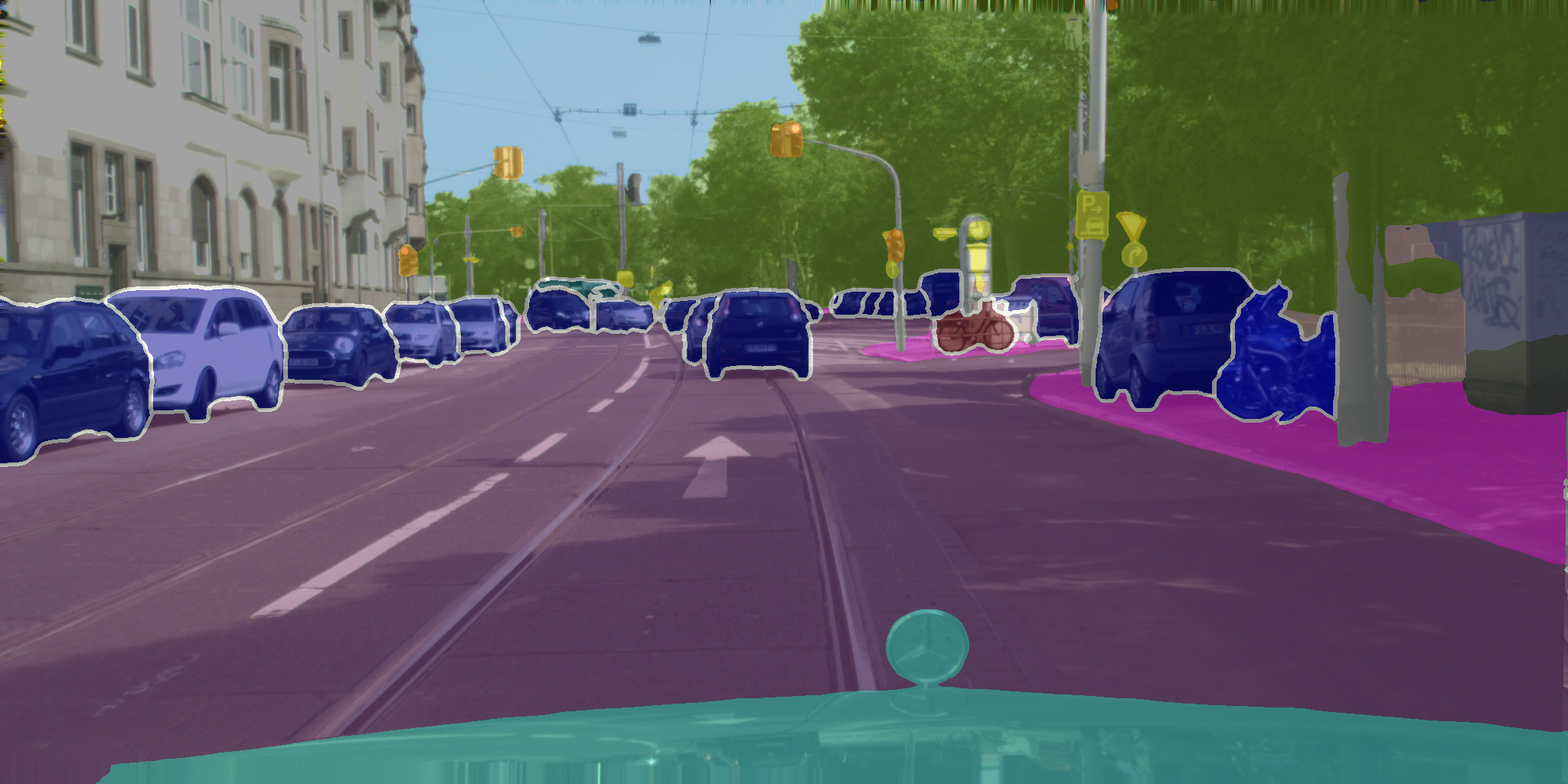}
  \caption{Panoptic segmentation}
  \label{fig:prediction:panoptic}
\end{subfigure}
\vspace*{\fill}
\begin{subfigure}[b]{0.48\linewidth}
  \centering
  \includegraphics[width=\linewidth]{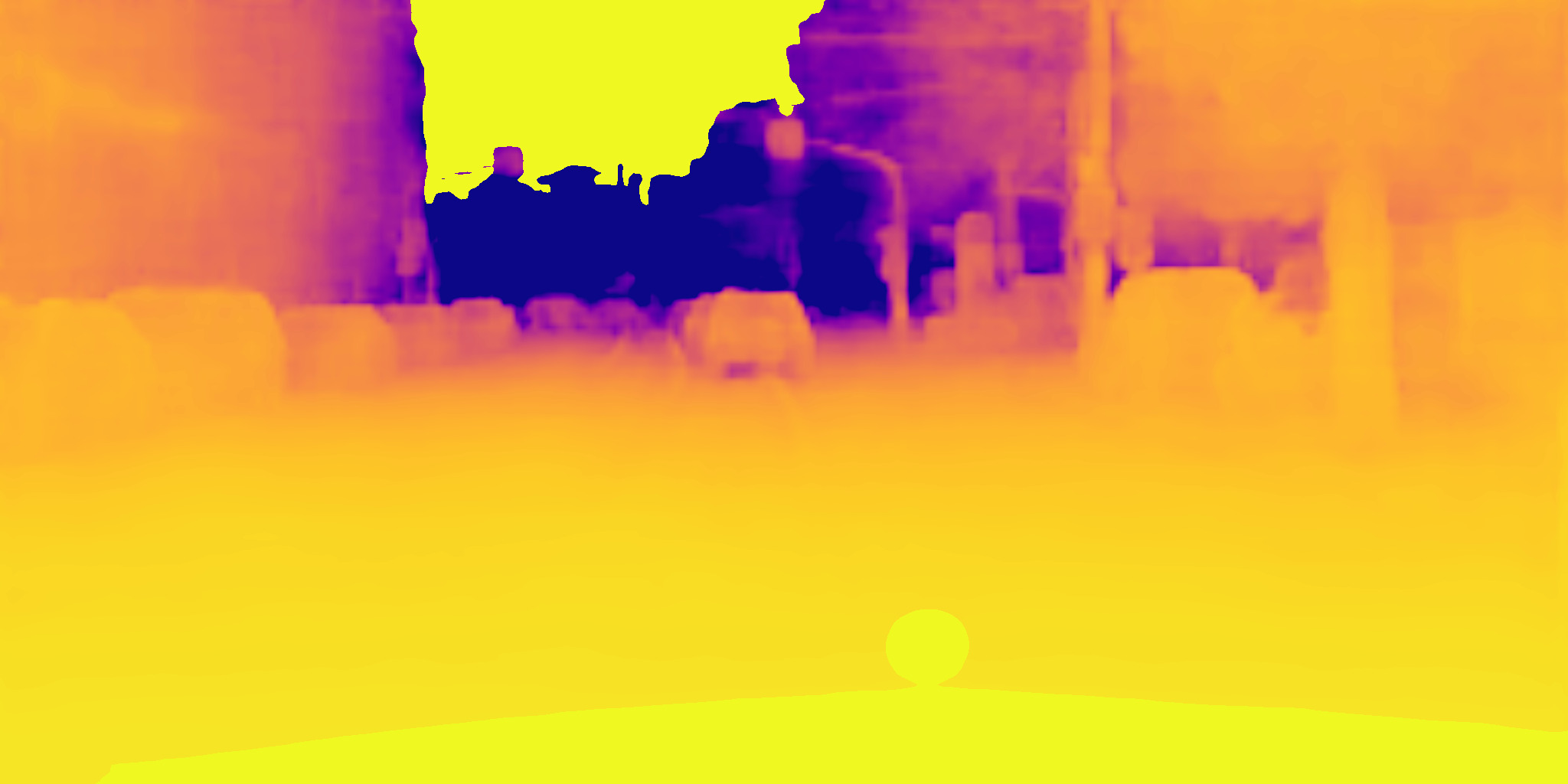}
  \caption{Monocular depth estimation}
  \label{fig:prediction:depth}
\end{subfigure}
\hspace*{\fill}
\begin{subfigure}[b]{0.48\linewidth}
  \centering
  \includegraphics[width=\linewidth]{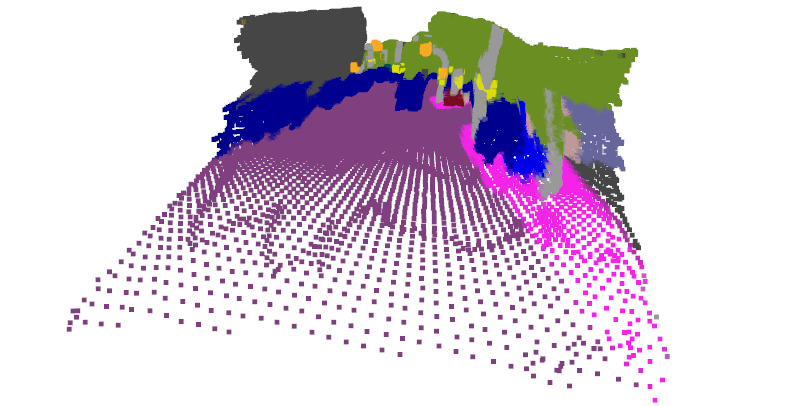}
  \caption{3D point cloud}
  \label{fig:prediction:pcl}
\end{subfigure}
\caption{Example prediction of our model, (\subref{fig:prediction:input}) the input image which is fed to the network, (\subref{fig:prediction:panoptic}) the panoptic prediction on top of the input image, (\subref{fig:prediction:depth}) the monocular depth estimation and (\subref{fig:prediction:pcl}) the final 3D point cloud generated by the network in real-time.}
\label{fig:prediction}
\end{figure}

\copyrightnotice
Scene understanding is an essential component for autonomous driving perception systems, since it provides necessary information for higher-level functions such as multi-object-tracking or behavior planning.
Recent advances in semantic segmentation~\cite{long2015fully, zhao2017pyramid, deeplabv3plus2018, WangSCJDZLMTWLX19, semantic_cvpr19, Tao2020HierarchicalMA} and instance segmentation~\cite{He_2017_ICCV, wang2020solo, wang2020solov2, homayounfar2020levelset, liang2020polytransform} based on deep neural networks show outstanding results, while fast models~\cite{yu2018bisenet, petrovai2020real, Hou_2020_CVPR, Gao2020LearningCA, DeGeus2020FastPanoptic} focus on optimizing the speed-accuracy trade-off for usage in latency-critical applications, \eg, autonomous driving.
New tasks emerge regularly, bringing perception systems a step forward towards full scene understanding, but also increasing task complexity.

One of these tasks is panoptic segmentation~\cite{kirillov2019panoptic}, a combination of semantic and instance segmentation.
Semantic segmentation focuses on stuff classes, \ie, amorph regions like sky or road, while instance segmentation focuses on thing classes, \ie, countable objects like cars, bicycles or pedestrians.
Panoptic segmentation handles both stuff and thing classes, providing not only unique class label for each pixel in the image, but also instance IDs for countable objects.
Panoptic segmentation is an important step towards scene understanding in autonomous vehicles, since it provides not only the object masks, but also interesting amorph regions like drivable road space or sidewalks.
However, the vast majority~\cite{xiong2019upsnet, Sofiiuk_2019_ICCV, li2018learning, Porzi_2019_CVPR, liu2019end, mohan2020efficientps, Qiao_2021_CVPR, Xiong_2019_CVPR, Porzi_2021_CVPR, Kirillov_2019_CVPR, Gao_2019_ICCV, yang2019deeperlab, cheng2020panoptic, chen2020naive, wang2020axial, wang2020max, li2021panopticfcn} of panoptic segmentation methods focus on high quality rather than inference speed, making them unsuitable for on-board integration into an autonomous vehicle perception system.
Only few methods~\cite{petrovai2020real, Hou_2020_CVPR, Gao2020LearningCA, DeGeus2020FastPanoptic} have been proposed in the low-latency regime.
Additionally, the camera-based measurement in the two-dimensional image plane limits the capabilities of panoptic segmentation models in autonomous vehicles.
The 2D pixel location is not sufficient for higher-level systems such as behavior planning to reason about the current environment. Rather, a 3D representation is required.

Monocular depth estimation~\cite{NIPS2014_7bccfde7} tackles this problem by predicting per-pixel depth estimates from single camera images.
Using this depth information, pixels from the image can be projected into 3D space.
Camera-based depth estimation is an ill-posed problem due to the camera measurement model, making it a very challenging task to solve.
Furthermore, accurate depth annotations are hard to acquire.
Stereo cameras can provide depth information, but require an accurate calibration between the two cameras.
Additionally, resulting depth images are noisy, inaccurate at farther distances, and have a lot of missing regions where the stereo matching failed.
In contrast, lidar sensors provide distance measurements with high accuracy.
The measurements can be projected into the image plane and used as a label, but this again requires an accurate calibration and synchronization between the sensors.
The generated depth images are much more accurate, but very sparse.
Therefore, recent methods follow a self-supervised training strategy using either stereo images~\cite{garg2016unsupervised, monodepth17}, video sequences~\cite{mahjourian2018unsupervised, zhou2019unsupervised, zhou2017unsupervised, zou2018df, casser2019depth, wang2018learning, gordon2019depth, packnet, zhou2020constant, xue2020toward}, or both~\cite{monodepth2, li2018undeepvo, luo2019every} during training.
The training objective is formulated as an image synthesis problem based on geometric constraints.

Combining multiple tasks can be resource demanding. 
The naive approach of training a separate network for each task can reach hardware limitations quickly in low resource environments.
Hence, multi-task learning~\cite{caruana1997multitask} emerged to combine multiple tasks in a single network to reduce latency and memory requirements on the target hardware.
While joint task training can potentially boost single-task performance, it brings its own difficulties, \eg, loss balancing and conflicting gradients.

In this work, we introduce the task of monocular geometric scene understanding, a combination of panoptic segmentation and self-supervised monocular depth estimation.
We propose a multi-task framework, which we call MGNet, to tackle this new task with focus on latency.
MGNet combines the ideas of state-of-the-art methods Panoptic Deeplab~\cite{cheng2020panoptic} and Monodepth2~\cite{monodepth2} with a lightweight network architecture.
The self-supervised monocular depth estimation formulation requires only video sequences during training, releasing us from the need of hard to acquire ground truth data.
Hence, our model can be trained using data from a single camera.
We propose an improved version of the \gls{dgc}, introduced in~\cite{xue2020toward}, using our panoptic prediction for scale-aware depth estimation.
Similar to~\cite{chen2020naive}, we generate pseudo labels for video sequence frames, which reduces the number of annotated frames necessary for panoptic segmentation.
The multi-task setting implicitly constrains the model to learn a unified representation for both tasks and reduces the overall latency.
We use homoscedastic uncertainty weighting, introduced in~\cite{kendall2018multi}, but adopt a novel weighting scheme, combining fixed and learnable task weights, to improve multi-task performance.
During deployment, our model produces 3D point clouds with instance aware class labels using a single camera image as input.
Figure~\ref{fig:prediction} shows an example prediction of our model.
We evaluate our method on Cityscapes~\cite{Cordts2016Cityscapes} and KITTI~\cite{Geiger2013IJRR} and are able to outperform previous approaches in terms of latency, while maintaining competitive accuracy.
Specifically, on Cityscapes, we achieve 55.7~PQ and 8.3~RMSE with 30~FPS on full resolution $1024\times2048$~pixel images.
On KITTI, we achieve 3.761~RMSE with 82~FPS on $384\times1280$~pixel images.
\section{Related Work}
\label{related_work}
Since, to the best of our knowledge, no combined algorithm exists yet, we discuss the related work separately for the fields of panoptic segmentation and self-supervised depth estimation.
Thereafter, we discuss related work in the field of multi-task learning for scene understanding.

\subsection{Panoptic Segmentation}
Panoptic segmentation~\cite{kirillov2019panoptic} was introduced to unify the tasks of semantic and instance segmentation.
Top-down approaches~\cite{li2018learning, xiong2019upsnet, Sofiiuk_2019_ICCV, Porzi_2019_CVPR, liu2019end, Qiao_2021_CVPR, Kirillov_2019_CVPR, Xiong_2019_CVPR, Chen_2020_CVPR, mohan2020efficientps, Porzi_2021_CVPR} first generate proposals, usually in form of bounding boxes, which are then used for instance mask prediction.
Most methods employ a Mask R-CNN~\cite{He_2017_ICCV} head combined with a separate branch for semantic segmentation and a merging module to handle conflicts~\cite{li2018learning, liu2019end, Porzi_2019_CVPR, mohan2020efficientps}.
For example, Mohan and Valada~\cite{mohan2020efficientps} propose a new variant of Mask R-CNN along with a new fusion module to achieve state-of-the-art results.
In contrast, bottom-up approaches~\cite{Gao_2019_ICCV, yang2019deeperlab, cheng2020panoptic, chen2020naive, wang2020axial, chen2020scaling} are proposal-free.
For example, Panoptic DeepLab~\cite{cheng2020panoptic} represents instance masks as pixel offsets and center keypoints.
An efficient merging module is employed to group the logits to final instance masks.
Previous works~\cite{chen2020naive, wang2020axial, QiaoVip_2021_CVPR, chen2020scaling} build on this simple and strong framework proving its flexibility.
Our work also uses the Panoptic DeepLab framework as a basis.
Recently, the first methods for end-to-end panoptic segmentation emerged~\cite{wang2020max, li2021panopticfcn}.
In contrast to previous works, these approaches predict panoptic segmentation maps directly using a unified representation for stuff and thing classes.
For example, Wang~\etal~\cite{wang2020max} introduced MaX-DeepLab, a novel architecture which extends Axial-DeepLab~\cite{wang2020axial} with a mask transformer head.
However, their method is computationally demanding, making it unsuitable for real-time applications.
Only a few works exist that have a focus on real-time panoptic segmentation~\cite{petrovai2020real, Hou_2020_CVPR, Gao2020LearningCA, DeGeus2020FastPanoptic}.
Hou~\etal~\cite{Hou_2020_CVPR} set state-of-the-art for real-time panoptic segmentation by proposing a novel panoptic segmentation network with efficient data flow, which can reach up to 30 FPS on full resolution images of Cityscapes.
Our model has similar performance while solving the additional task of self-supervised monocular depth estimation.

\subsection{Self-Supervised Monocular Depth Estimation}
Zhou~\etal~\cite{zhou2017unsupervised} were the first to propose a training scheme for depth estimation that only uses monocular video sequences for supervision.
The idea is to synthesize the current frame from adjacent frames in the video sequence by using the predicted depth estimation and predicted relative pose between the frames.
The synthesized frames are then compared with the current frame using a photometric loss.
During training, multiple depth maps at different scales are used to mitigate the effect of learning from low texture regions.
Assumptions for self-supervised depth learning are a moving camera in a static environment and no occlusions between the frames.
Furthermore, due to the ambiguous nature of photometric loss, the depth can only be predicted up to an unknown scale factor.
Since then, a number of works~\cite{mahjourian2018unsupervised, zhou2019unsupervised, zou2018df, casser2019depth, wang2018learning, gordon2019depth, packnet, zhou2020constant, xue2020toward, monodepth2, li2018undeepvo, luo2019every} advanced the field considerably.
For example, Godard~\etal~\cite{monodepth2} propose to upsample the multi-scale depth maps before loss calculation and use the minimum photometric error to tackle occlusions.
Furthermore, they propose an auto-masking strategy to avoid holes of infinite depth in low-texture regions and in regions with dynamic objects.
Xue~\etal~\cite{xue2020toward} build on top of~\cite{monodepth2} and propose a \gls{dgc} to estimate the scale factor based on geometric constraints.
We incorporate the basic idea of~\cite{xue2020toward} in our framework, but introduce an improved version of the \gls{dgc} which leverages our panoptic prediction.
Some works focus on incorporating semantic~\cite{packnet-semguided, klingner2020self} or panoptic~\cite{saeedan2021boosting} label maps to explicitly boost the performance of self-supervised depth maps.
In contrast, our method uses multi-task optimization to implicitly boost single-task performance.

\subsection{Multi-Task Learning}
Multi-Task learning~\cite{caruana1997multitask} has emerged in order to save computational resources by combining multiple tasks in a single network.
Furthermore, learning multiple tasks at once can improve generalization ability and lead to better results compared to single-task performance.
A number of works exists, which tackle the different tasks for scene understanding in a multi-task setting~\cite{wang2020sdc, neven2017fast, zamir2020robust, vandenhende2020mti, zhang2019pattern, xu2018pad, jha2020adamt, kendall2018multi, QiaoVip_2021_CVPR, goel2021quadronet}.
Goel~\etal~\cite{goel2021quadronet} propose QuadroNet, a real-time capable model to predict 2D bounding boxes, panoptic segmentation, and depth from single images.
While similar to our work with respect to prediction, they train depth in a fully supervised way, requiring disparity or lidar ground truth during training.
Some works incorporate semantic segmentation and self-supervised depth estimation in multi-task frameworks~\cite{klingner2020improved, kumar2021syndistnet}.
Klingner~\etal~\cite{klingner2020improved} show that joint training of self-supervised depth and supervised semantic segmentation can boost performance and increase noise robustness of the model.
Our model does not combine semantic segmentation, but rather panoptic segmentation with self-supervised depth estimation.

\section{Method}
\label{method}

\begin{figure*}[t]
\includegraphics[width=\textwidth]{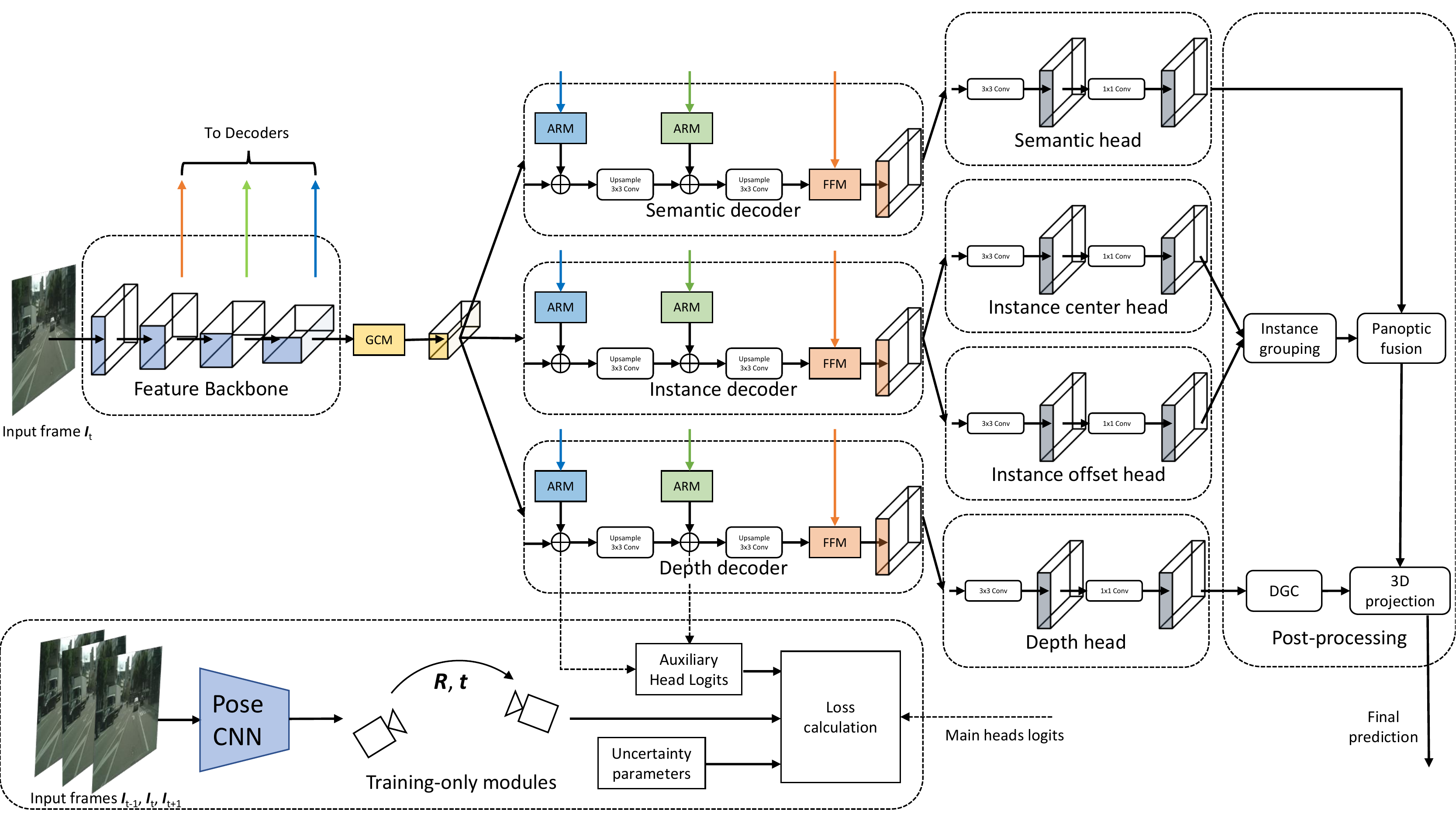}
\caption{Overview of our MGNet framework. We encode image features using a lightweight backbone and employ a \gls{gcm} to maximize the receptive field. Three task-specific decoders are used to fuse backbone features of different scales using \glspl{arm} and \glspl{ffm}. Task specific heads convert the fused features to logits. Object instances are represented as instance centers and pixel-wise offsets, i.e., 2D vectors, representing the x-y-offset to the instance center. The post-processing groups instances to the closest center given the offset prediction and assigns a semantic class based on majority voting. The depth prediction is first scaled using a \acrfull{dgc} and then used to convert the panoptic prediction to the final 3D point cloud prediction. The pose network predicting 6 DOF poses between adjacent camera frames $\vm{I}_{t-1}, \vm{I}_{t}$ and $\vm{I}_{t+1}$, auxiliary heads for the multi-view photometric loss, and the uncertainty parameters for multi-task optimization are only used during training.}
\label{fig:overview}
\end{figure*}

In this section, we describe our MGNet framework to jointly tackle the tasks of panoptic segmentation and self-supervised monocular depth estimation in a single efficient model.
Figure~\ref{fig:overview} gives an overview of our framework.
The following sections explain the different parts of our framework in detail.

\subsection{MGNet Network Architecture}

Our MGNet architecture is designed as an encoder-decoder structure with a single shared encoder and three task-specific decoders.

\textbf{Encoder:} The encoder consists of a feature extraction backbone, which extracts image level features at different scales up to an output stride of 32.
The backbone can easily be switched, hence we compare different lightweight backbones in Section \ref{experimental_results} with respect to overall performance and latency.
Additionally, we add a \acrfull{gcm} on top of the last feature map of the backbone.
The \gls{gcm} uses global average pooling to extract fine-grained features from the input image.

\textbf{Task-specific decoders:} The task-specific decoders all share the same structure.
For panoptic segmentation, we use two decoders as proposed in~\cite{cheng2020panoptic}, one for semantic segmentation and one for instance segmentation.
A single decoder is used for the task of self-supervised depth estimation.
We leverage \acrfullpl{arm} and \acrfullpl{ffm} from~\cite{yu2018bisenet} to combine the features from different scales in an efficient and effective way.
Similar to~\cite{yu2018bisenet}, we use two \glspl{arm} on the last two feature maps of the backbone and add them together with the fine-grained features generated by the global context module.
In contrast to~\cite{yu2018bisenet}, we do not follow the bilateral feature encoding using a spatial path for low-level and a context path for high-level features.
We found that the bilateral approach does not improve performance, but increases latency of the overall model.
Instead, we use a skip connection to propagate low-level features from the backbone to the decoders and combine them with the high-level features in the \gls{ffm}.

\textbf{Head modules:} We add simple head modules to each decoder to generate the logits.
All heads share the same structure, using a $3\times3$~conv followed by a $1\times1$~conv to map the feature channel size to the final logits channel size, \eg, the number of semantic classes used for semantic segmentation.
Similar to~\cite{cheng2020panoptic}, we encode instances by their center of mass and pixel-wise 2D offset vectors pointing towards the center.
We use a separate head for instance center and offset regression, but share the decoder for the instance task.
The logits produced by the depth head are passed through a sigmoid activation as used in~\cite{monodepth2}.

\subsection{Multi-Task Optimization}
\label{multi_task_optim}
\textbf{Panoptic Segmentation:} We follow the Panoptic DeepLab approach~\cite{cheng2020panoptic} and use the weighted bootstrapped cross entropy loss, first introduced in~\cite{yang2019deeperlab}, for semantic segmentation
\begin{equation}
\mathcal{L}_{\text{seg}} = -\frac{1}{K}\sum_{i=1}^{N}\omega_i\cdot\mathds{1}\left[p_{i,y_i}<t_K\right]\cdot\log p_{i,y_i},
\end{equation}
where $p_{i,y_i}$ is the predicted probability for pixel $i$ and target class $y_i$.
The indicator function $\mathds{1}\left[p_{i,y_i}<t_K\right]$ ensures that the loss is masked using only pixels with the top-K highest loss values.
The weight is set to $w_i=3$ for pixels that belong to instances with a smaller area than $64\times64$ and $w_i=1$ for all other pixels.
This way, the network is forced to focus on hard pixels and small instances, respectively.
For the instance segmentation, we represent each instance by its center of mass.
During training, we predict a 2D Gaussian heatmap for each instance center with a fixed sigma of 8 pixels and use a Mean Squared Error (MSE) loss for optimization.
For the instance mask prediction, we further encode pixels which belong to thing classes by their offset vector to the corresponding instance center.
We use a $L_1$ loss for optimization of the offset vectors prediction.
We weight each loss term according to~\cite{yang2019deeperlab}, hence the full panoptic segmentation loss is given by
\begin{equation}
\mathcal{L}_{\text{pan}} = \mathcal{L}_{\text{seg}} + 200\cdot \mathcal{L}_{\text{mse}} + 0.01\cdot \mathcal{L}_{L_1}.
\end{equation}

\textbf{Self-Supervised Monocular Depth Estimation:} We follow recent methods~\cite{monodepth2, packnet}, and use a multi-scale photometric loss.
We add two auxiliary heads to produce intermediate depth logits.
During training, we use a separate pose network based on ResNet18~\cite{he2016deep} to predict the 6 DOF relative camera pose between the adjacent frames $\vm{I}_{t-1}, \vm{I}_t$ and $\vm{I}_{t+1}$, and optimize both networks together.
For loss calculation, the context frames $\vm{I}_c\in\{\vm{I}_{t-1}, \vm{I}_{t+1}\}$ are first warped into the frame $\vm{I}_{t}$ using the predicted depth maps and poses.
For each warped image $\hat{\vm{I}}_t\in\{\hat{\vm{I}}_{t-1\rightarrow t}, \hat{\vm{I}}_{t+1\rightarrow t}\}$, the photometric loss is calculated according to
\begin{equation}
\mathcal{L}_{\text{phot}}(\vm{I}_{t}, \hat{\vm{I}}_t) = \alpha \cdot \frac{1-\text{SSIM}(\vm{I}_{t}, \hat{\vm{I}}_t)}{2} + (1-\alpha)\cdot |\vm{I}_{t}-\hat{\vm{I}}_t|
\end{equation}
using the Structured Similarity Index (SSIM)~\cite{wang2004image}.
We use a fixed $\alpha=0.85$ in this work.
Following~\cite{monodepth2}, we integrate several improvements into the photometric loss calculation.
Instead of averaging over the photometric loss terms for each context frame, we use the minimum reprojection error
\begin{equation}
\mathcal{L}_{\text{phot}}(\vm{I}_{t}, \hat{\vm{I}}_s) = \min_{\hat{\vm{I}}_s}\mathcal{L}_{\text{phot}}(\vm{I}_{t}, \hat{\vm{I}}_s)
\end{equation}
with $\hat{\vm{I}}_s\in\{\hat{\vm{I}}_{t-1\rightarrow t}, \hat{\vm{I}}_{t+1\rightarrow t},\vm{I}_{t-1}, \vm{I}_{t+1}\}$.
Adding the unwarped image loss terms $\mathcal{L}_{\text{phot}}(\vm{I}_{t},\vm{I}_{t-1})$ and $\mathcal{L}_{\text{phot}}(\vm{I}_{t},\vm{I}_{t+1})$ automatically masks out static pixels from the loss.
Furthermore, we use a pixel-wise mask $\vm{M}_v$ to mask out pixels without a valid projection.
The final photometric loss thus is given by
\begin{equation}
\mathcal{L}_{\text{phot}} = \sum_i{\mathcal{L}_{\text{phot}}(\vm{I}_{t}, \vm{I}_c)\odot\vm{M}_v},
\end{equation}
where $i$ denotes predictions at different scales.
Similar to~\cite{monodepth17}, we regularize the predicted depth map using an image gradient based smoothness term
\begin{align}
\begin{split}
\mathcal{L}_{\text{smooth}} = \sum_i\tfrac{1}{2^i}(&|\delta_x\hat{\vm{d}}^*|\exp(-|\delta_x\vm{I}_{t}|)\\ 
+ &|\delta_y\hat{\vm{d}}^*|\exp(-|\delta_y\vm{I}_{t}|)).
\end{split}
\end{align}
with the predicted mean-normalized inverse depth $\hat{\vm{d}}^*$.
The final loss term for self-supervised depth estimation is a weighted sum of the photometric and smoothness loss
\begin{equation}
\mathcal{L}_{\text{depth}} = \mathcal{L}_{\text{phot}} + 0.001\cdot\mathcal{L}_{\text{smooth}}.
\end{equation}

\textbf{Homoscedastic uncertainty weighting:} The naive approach of adding up both loss terms may be suboptimal for optimization.
Hence, we use the homoscedastic uncertainty~$\sigma_i^2$ to weight each task~$i$ as proposed in~\cite{kendall2018multi}.
We add uncertainty terms not only to the final task losses, but to each component of each loss.
We argue that the fixed loss weights give a good first estimate for optimization, although this might change during training.
Therefore, including learnable uncertainties enables the model to adjust the individual loss term relations based on training data.
In practice, we learn $s_i=\log(\sigma_i^2)$ instead of $\sigma_i^2$, leading to the final loss
\begin{equation}
\begin{split}
\mathcal{L} &= \exp(-s_0)\cdot \mathcal{L}_{\text{seg}} + 0.5\cdot\exp(-s_1)\cdot 200.\cdot \mathcal{L}_{\text{mse}}  \\
&+ 0.5\cdot\exp(-s_2)\cdot 0.01\cdot \mathcal{L}_{L_1} + 0.5\cdot\exp(-s_3)\cdot \mathcal{L}_{\text{phot}} \\
&+ 0.5\cdot\exp(-s_4)\cdot 0.001\cdot\mathcal{L}_{\text{smooth}} + 0.5\cdot\sum_i s_i.
\end{split}
\end{equation}

\subsection{Post-Processing}
The post-processing includes instance grouping, panoptic fusion, depth scaling, and 3D projection.

\textbf{Instance grouping:} Center keypoints are extracted from the heatmap prediction using a keypoint-based non-maximum suppression (NMS) as proposed in~\cite{cheng2020panoptic}.
In this method, max pooling with kernel size 7 is applied to the heatmap prediction and pixels are kept only if the pooled prediction is equal to the unpooled prediction.
Additionally, a fixed threshold of 0.3 is used to filter out low confidence keypoints.
The offset prediction is masked with the semantic prediction to only preserve pixels belonging to thing classes.
Instance masks can then be generated by grouping the thing pixels to their closest center keypoint after applying the predicted offsets.

\textbf{Panoptic fusion:} Panoptic fusion assigns class labels to the grouped instances based on the semantic prediction.
We use the simple and efficient majority voting introduced in~\cite{yang2019deeperlab} to assign to each instance the class label with the highest pixel count for that mask.

\textbf{Depth scaling:} The depth estimation can only be done up to an unknown scale factor due to the inherent ambiguity of the photometric loss.
To rescale the estimated depth map $\vm{d}_{\text{rel}}$ back to the original scale, we leverage the \gls{dgc} from~\cite{xue2020toward}, but use our panoptic prediction to improve the module.
The idea is to estimate a camera height for each ground point based on the predicted depth using the camera geometry.
In the origial \gls{dgc}, $\vm{d}_{\text{rel}}$ is used to generate 3D points $\vm{p}_{i}=(x,y,z)^\intercal$ from the image and a surface normal $\vm{N}(\vm{p}_{i})$ is calculated for each 3D point to classify points as ground points, where the surface normal is close to the ideal ground surface normal $\vm{n}=(0,1,0)^\intercal$.
However, this approach produces lots of false positives, \eg, roofs of vehicles.
To mitigate this, we use our panoptic prediction, \ie, points with class label \texttt{road}, for ground point classification instead.
Using these ground points, we estimate a camera height for each ground point using
\begin{equation}
h(\vm{p}_{i})=\vm{N}(\vm{p}_{i})^\intercal\vm{p}_{i}.
\end{equation}
The scale factor is then calculated using the real camera height $h$ and the median of the estimated camera heights $\hat{h}_{\text{m}}$. Hence, the scaled depth prediction can be calculated as
\begin{equation}
\vm{d}_{\text{abs}} = \frac{h}{\hat{h}_{\text{m}}}\cdot \vm{d}_{\text{rel}}.
\end{equation}
 
\textbf{3D projection:} The 3D projection of the panoptic prediction $\hat{\vm{y}}(u, v)$ is calculated by
\begin{equation}
\hat{\vm{y}}(x, y, z)=\vm{K}^{-1}\vm{d}_{\text{abs}}\hat{\vm{y}}(u, v)
\end{equation}
using the scaled depth prediction $\vm{d}_{\text{abs}}$ and the intrinsic camera matrix $\vm{K}$.
In order to only generate 3D points with a valid depth prediction, we exclude certain semantic classes, in our case \texttt{sky} and \texttt{ego-car}, from the projection.
\section{Experimental Results}
\label{experimental_results}

In this section, we evaluate our multi-task framework quantitatively and qualitatively and provide comparisons with state-of-the-art methods of both domains, panoptic segmentation and self-supervised depth estimation.

\subsection{Datasets}
We conduct experiments on two popular autonomous driving benchmarks, Cityscapes~\cite{Cordts2016Cityscapes} and KITTI~\cite{Geiger2013IJRR}.

\textbf{Cityscapes:} The Cityscapes dataset provides video sequences recorded in 50 different cities primarily in Germany.
It contains 5\,000 images that are annotated with fine-grained panoptic labels for 30 classes, of which 19 are used during evaluation.
The dataset is split in 2\,975 images for training, 500 for validation and 1\,525 for testing.
Additional 20\,000 images provide coarse panoptic labels, which are not used in this work.
Depth annotations are given as pre-computed disparity maps from SGM~\cite{hirschmuller2007stereo} for the 5\,000 fine annotated images, which are only used for evaluation.
We train with 20 classes, including the 19 classes used for evaluation and the additional class \texttt{ego-car}.
We found that using this extra class improves panoptic prediction and enables us to exclude pixels belonging to this class from photometric loss calculation.

\textbf{KITTI Eigen split:} We use the KITTI Eigen split introduced in~\cite{NIPS2014_7bccfde7} with pre-processing from Zhou~\etal~\cite{zhou2017unsupervised} to remove static frames.
This leads to an image split of 39\,810 images for training and 4424 for validation.
We use the 652 improved ground truth depth maps introduced in~\cite{Uhrig2017THREEDV} for testing.
The depth maps are generated using accumulated lidar points from 5 consecutive frames and stereo information to handle dynamic objects.
KITTI does not provide panoptic segmentation annotations for the Eigen split.
Hence, we use our best Cityscapes model to generate pseudo labels as in~\cite{chen2020naive} and train on pseudo labels only.
We exclude the \texttt{ego-car} class, as KITTI images do not have a visible part of the ego vehicle.

\subsection{Implementation Details}
We use the PyTorch framework~\cite{NEURIPS2019_9015} and train all models across 4 NVIDIA RTX 2080Ti GPUs.
We use InPlaceABNSync~\cite{rotabulo2017place} with LeakyReLU~\cite{maas2013rectifier} activation after each convolution except the last ones in the heads.
We adopt a similar training protocol as in~\cite{cheng2020panoptic}.
In particular, we use the Adam Optimizer~\cite{kingma2014adam} without weight decay and adopt the ``poly'' learning rate schedule~\cite{liu2015parsenet} with an initial learning rate of 0.001.
We multiply the learning rate by a factor of 10 in all our decoders and heads.
Our model backbone as well as the pose network are initialized with pre-trained weights from ImageNet~\cite{deng2009imagenet}.
We do not use other external data such as Mapillary Vistas~\cite{MVD2017}.
During training, we do random color jitter augmentation.
Additionally, on Cityscapes, we perform random scaling within [0.5, 2.0] and, respectively, randomly crop or pad the images to a fixed size of $1024\times1024$ pixels.
We adjust the intrinsic camera matrix according to the augmentation and calculate a reprojection mask to exclude padded regions as well as \texttt{sky} and \texttt{ego-car} pixels from the photometric loss calculation.
Similar to~\cite{chen2020naive}, we use video sequences to generate pseudo labels for training, which enables us to use the full video sequence set for a second training iteration.
On Cityscapes, we train for 60k iterations on the fine-annotated set and another 60k iterations on the video sequence set using pseudo labels with batch size 12.
On KITTI, we use our best Cityscapes model to generate pseudo labels for panoptic segmentation and train our model for 20 epochs with batch size 12.

\begin{table}
\begin{center}
\begin{tabular}{|l|c|c|c|c|}
\hline
Backbone & PQ $\uparrow$ & RMSE $\downarrow$ & FPS $\uparrow$ \\
\hline\hline
MobileNetV3~\cite{howard2019searching} & 48.6 & 9.4 & 29.8 \\
MNASNet100~\cite{tan2019mnasnet} & 50.8 & 9.0 & 28.3 \\
EfficientNetLite0~\cite{tan2019efficientnet} & 52.8 & \textbf{8.6}  & 27.7 \\
ResNet18~\cite{he2016deep} & \textbf{53.3} & 8.8 & \textbf{30.0} \\
\hline
\end{tabular}
\end{center}
\caption{Comparison of different feature backbones. We compare four lightweight ImageNet~\cite{deng2009imagenet} pre-trained backbones.}
\label{tab:backbone_comp}
\end{table}

\begin{table}
\begin{center}
\begin{tabular}{|l|c|c|c|c|c|c|c|c|}
\hline
Method & PQ $\uparrow$ & RMSE $\downarrow$ \\
\hline\hline
Single-Task Baseline & 54.1 & 9.3 \\
Multi-Task Baseline & 53.3 & 8.8 \\
+ Uncertainty & 54.2 & 8.6 \\
+ Video Sequence~\cite{chen2020naive} & \textbf{55.7} & \textbf{8.3} \\
\hline
\end{tabular}
\end{center}
\caption{Ablation study on Cityscapes. While PQ decreases in the multi-task baseline compared to two single-task models, adding uncertainty weighting and video sequence training show improvements in both, PQ and RMSE.}
\label{tab:ablation}
\end{table}

In order to evaluate the panoptic segmentation task, we report the panoptic quality (PQ), introduced in~\cite{kirillov2019panoptic}, where a higher PQ states a more accurate prediction. 
RMSE is used as the main metric for the evaluation of self-supervised depth estimation.
Furthermore, we report other common depth metrics, namely absolute relative error and accuracy under delta threshold for $\delta<1.25, \delta<1.25^2$, and $\delta<1.25^3$.
In case of RMSE and absolute relative error, a lower value is better, while higher values are better in case of the delta threshold metrics.
Additionally, we measure 500 forward passes through our network, including data loading to GPU and post-processing, and report the average runtime or FPS, respectively, to assess the real-time capability of the models.
We use TensorRT optimization~\cite{tensorrt} unless stated otherwise.

\begin{table*}
\begin{center}
\begin{tabular}{|l|c|c|c|c|c|}
\hline
Method & Backbone & Input size & PQ $\uparrow$ & GPU &  Runtime (ms) $\downarrow$ \\
\hline\hline
Naive Student~\cite{chen2020naive}\dag & WideResNet41 & $1025\times2049$ & \textbf{70.8} & V100 & 396.5 \\
Porzi~\etal~\cite{Porzi_2021_CVPR}*\dag & HRNet-W48+ & $1024\times2048$ & 66.7 & V100 & $-$ \\
Axial-DeepLab-XL~\cite{wang2020axial}* & Axial-ResNet-XL & $1025\times2049$ & 65.1 & V100 & $-$ \\
Panoptic DeepLab~\cite{cheng2020panoptic} & Xception71 & $1025\times2049$ & 63.0 & V100 & 175 \\
EfficientPS~\cite{mohan2020efficientps} & Mod. EfficientNetB5 & $1024\times2048$ & 63.6 & Titan RTX & 166 \\
Panoptic FCN~\cite{li2021panopticfcn}* & Res50-FPN & $1024\times2048$ & 61.4 & V100 & $-$ \\
FPSNet~\cite{DeGeus2020FastPanoptic} & ResNet50-FPN & $1024\times2048$ & 55.1 & Titan RTX & 114 \\
Hou~\etal~\cite{Hou_2020_CVPR} & ResNet50-FPN & $1024\times2048$ & 58.8 & V100 & 99 \\
Petrovai and Nedevschi~\cite{petrovai2020real} & VoVNet2-39 & $1024\times2048$ & 63.0 & V100 & 82 \\
Panoptic DeepLab~\cite{chen2020scaling} & SWideRNet-(0.25, 0.25, 0.75) & $1025\times2049$ & 58.4 & V100 & 63.05 \\
Panoptic DeepLab~\cite{cheng2020panoptic} & MobileNetV3 & $1025\times2049$ & 55.4 & V100 & 63 \\
\hline
\textbf{Ours} & ResNet18 & $1024\times2048$ & 55.7 & RTX 2080 Ti & \textbf{44.4} \\
\hline
\end{tabular}
\end{center}
\caption{Comparison of our method to state-of-the-art for the panoptic segmentation task on the Cityscapes validation set. Methods marked with * do not report runtimes, but based on model size we estimate $>$100~\si{ms}. Methods using external data, \ie, Mapillary Vistas~\cite{MVD2017}, are marked with\,\dag.}
\label{tab:cityscapes}
\end{table*}

\begin{table*}
\begin{center}
\begin{tabular}{|l|c|c|c|c|c|c|c|c|c|}
\hline
Method & DS & Resolution & Abs Rel $\downarrow$ & RMSE $\downarrow$ & $\delta<1.25 \uparrow$ & $\delta<1.25^2 \uparrow$ & $\delta<1.25^3 \uparrow$ \\
\hline\hline
SfMLearner~\cite{zhou2017unsupervised} & CS+K & $416\times128$ & 0.176 & 6.129 & 0.758 & 0.921 & 0.971\\
Vid2Depth~\cite{mahjourian2018unsupervised} & CS+K & $416\times128$ & 0.134 & 5.501 & 0.827 & 0.944 & 0.981\\
GeoNet~\cite{yin2018geonet} & CS+K & $416\times128$ & 0.132 & 5.240 & 0.883 & 0.953 & 0.985\\
DDVO~\cite{wang2018learning} & CS+K & $416\times128$ & 0.126 & 4.932 & 0.851 & 0.958 & 0.986\\
EPC++~\cite{luo2019every} & K & $640\times192$ & 0.120 & 4.755 & 0.856 & 0.961 & 0.987\\
Monodepth2~\cite{monodepth2} & K & $640\times192$ & 0.090 & 3.942 & 0.914 & 0.983 & 0.995\\
SynDistNet~\cite{kumar2021syndistnet} & K & $640\times192$ & 0.076 & 3.406 & 0.931 & 0.988 & 0.996\\
PackNet-SfM~\cite{packnet} & CS+K & $1280\times384$ & \textbf{0.071} & \textbf{3.153} & \textbf{0.944} & \textbf{0.990} & \textbf{0.997}\\
\hline
\textbf{Ours} & CS+K & $1280\times384$ & 0.095 & 3.761 & 0.902 & 0.979 & 0.992\\
\hline
\end{tabular}
\end{center}
\caption{Comparison of our method to state-of-the-art self-supervised depth estimation methods on the KITTI 2015 Eigen split for a distance up to 80~\si{m}. We use the improved ground truth maps from~\cite{Uhrig2017THREEDV} for comparison.}
\label{tab:kitti}
\end{table*}

\subsection{Ablation Studies}
We perform ablation studies on Cityscapes.
First, we compare four lightweight backbones, namely MobileNetV3~\cite{howard2019searching}, MNASNet100~\cite{tan2019mnasnet}, EfficientNetLite0~\cite{tan2019efficientnet} and ResNet18~\cite{he2016deep}.
The results are reported in Table~\ref{tab:backbone_comp}.
ResNet18 is performing best in terms of PQ and FPS, while being only slightly worse in terms of RMSE compared to EfficientNetLite0.
MobileNetV3 and MNASNet100 perform significantly worse.
Hence, we use ResNet18 as backbone for all further experiments.
Additionally, we investigate the effect of multi-task training, uncertainty weighting, and video sequence training in Table~\ref{tab:ablation}.
The naive multi-task setting with fixed weights improves RMSE, but deteriorates PQ compared to two seperate single-task models.
However, adding uncertainty weighting, we see an improvement of 0.1\% PQ and 0.7~\si{m} RMSE compared to the single-task baseline.
By adding video sequence training, we can further increase performance to 55.7~PQ and 8.3~RMSE.

\begin{figure*}[t]
\begin{subfigure}[b]{0.24\textwidth}
  \centering
  \includegraphics[width=\textwidth]{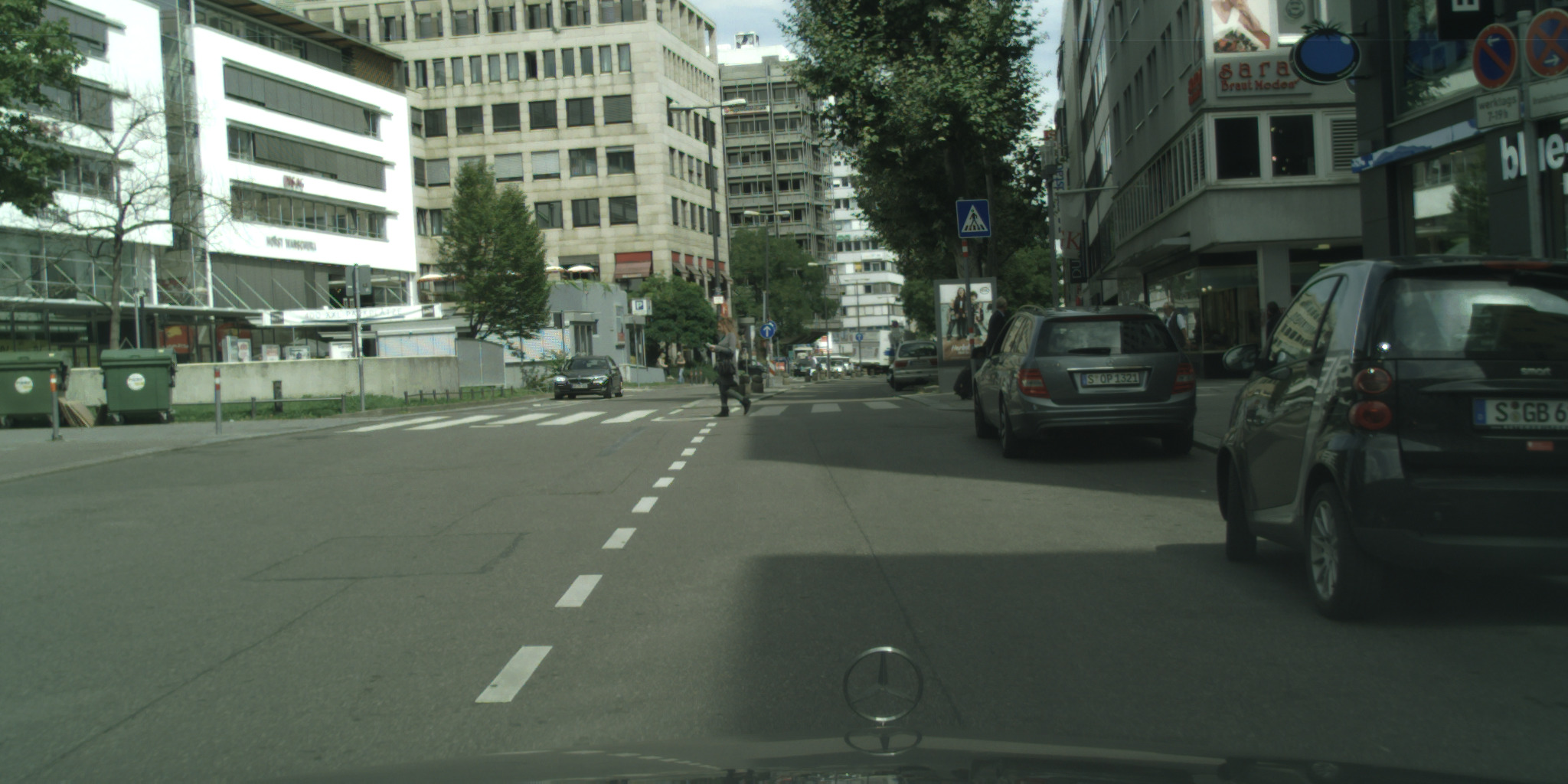}
\end{subfigure}
\hspace*{\fill}
\begin{subfigure}[b]{0.24\textwidth}
  \centering
  \includegraphics[width=\textwidth]{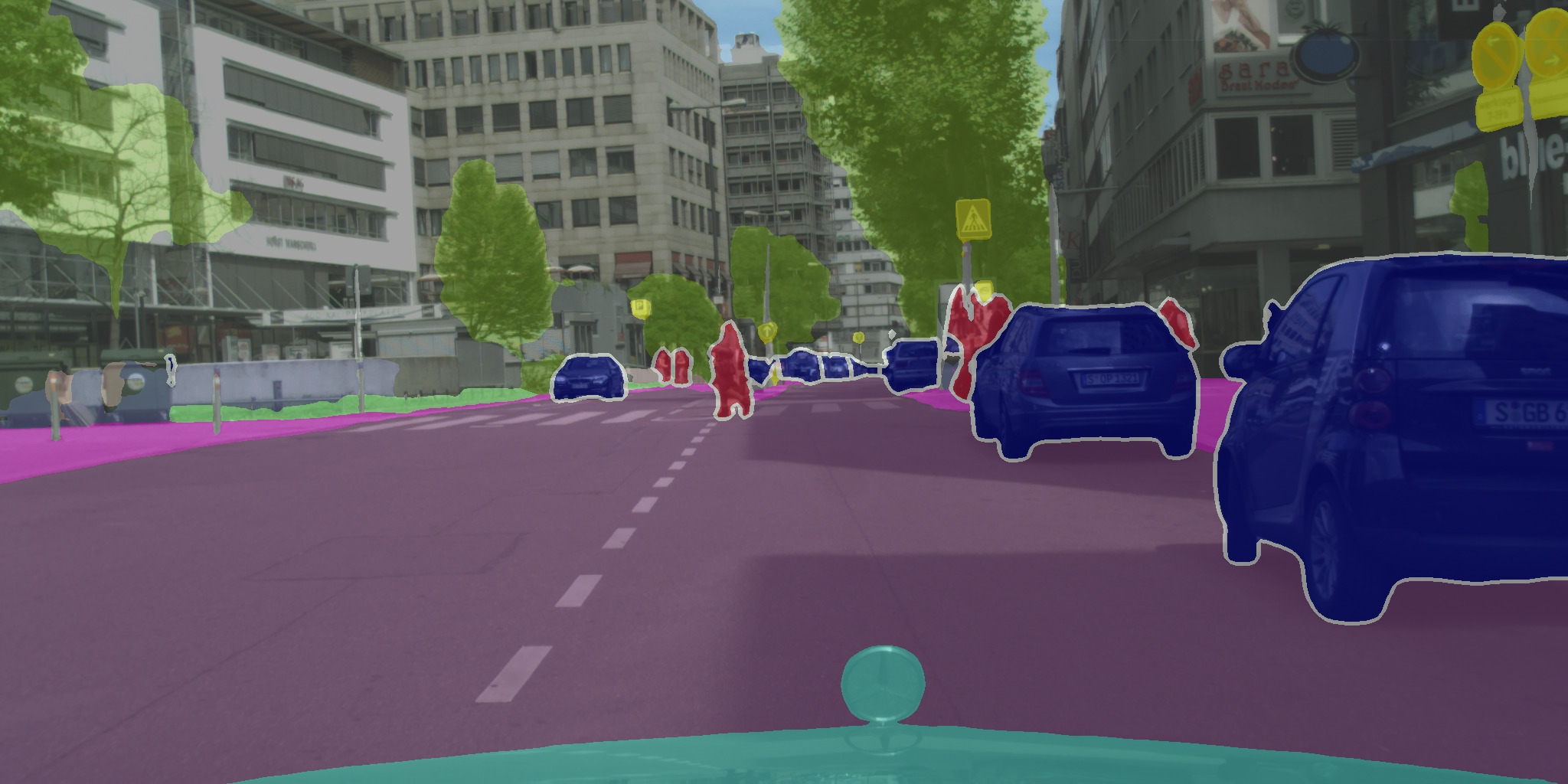}
\end{subfigure}
\begin{subfigure}[b]{0.24\textwidth}
  \centering
  \includegraphics[width=\textwidth]{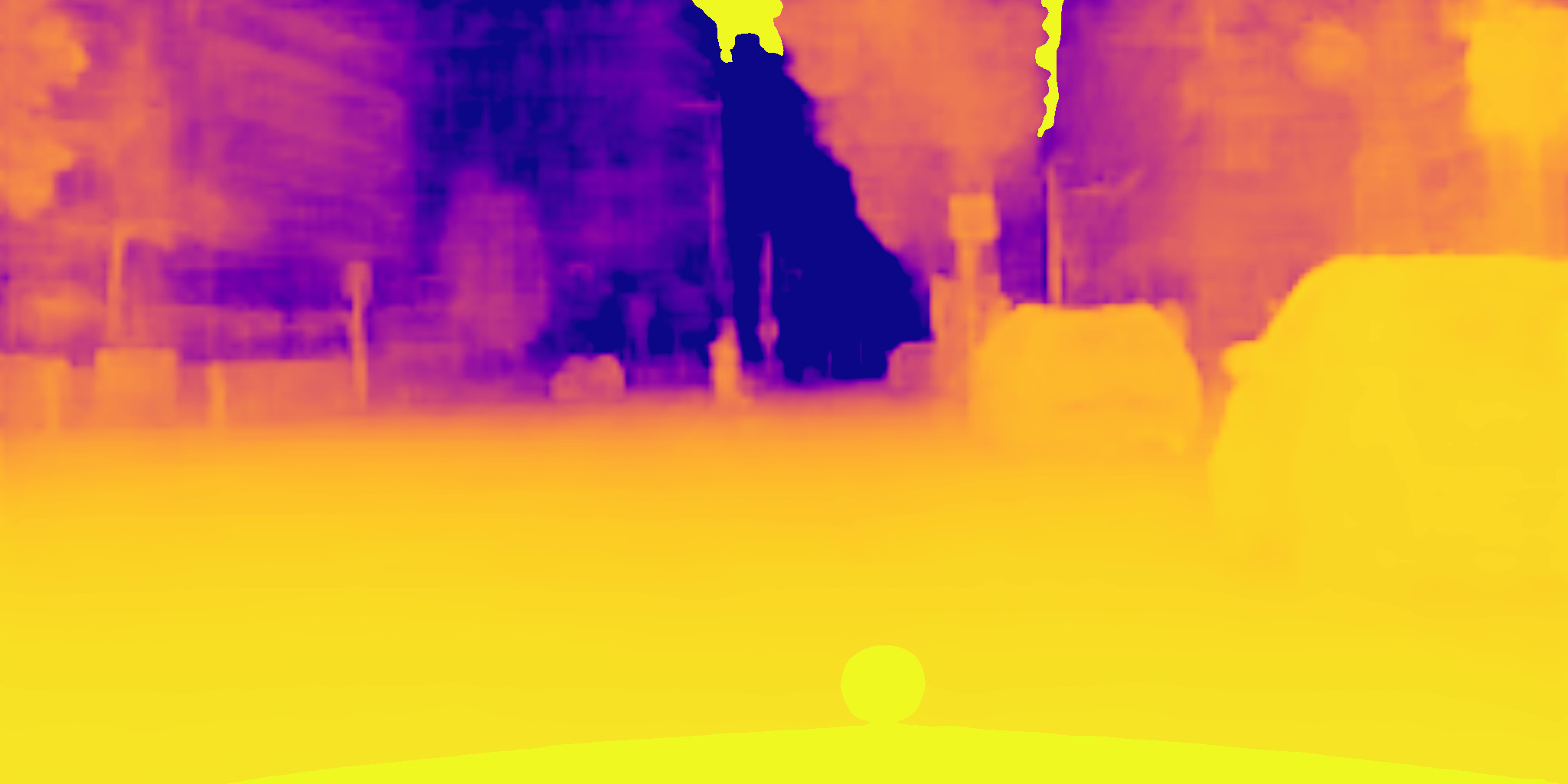}
\end{subfigure}
\hspace*{\fill}
\begin{subfigure}[b]{0.24\textwidth}
  \centering
  \includegraphics[width=\textwidth]{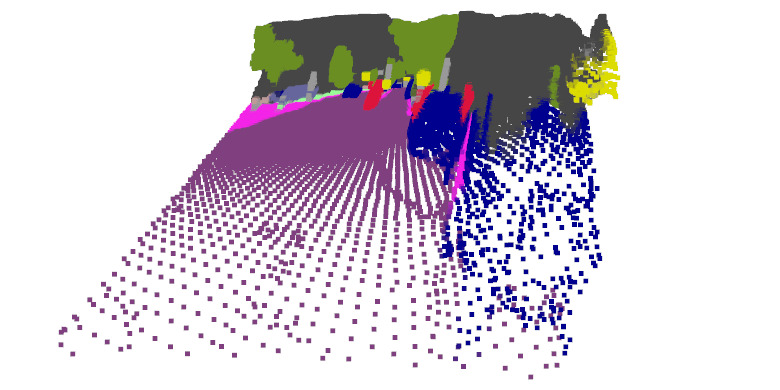}
\end{subfigure}
\vspace*{\fill}
\begin{subfigure}[b]{0.24\textwidth}
  \centering
  \includegraphics[width=\textwidth]{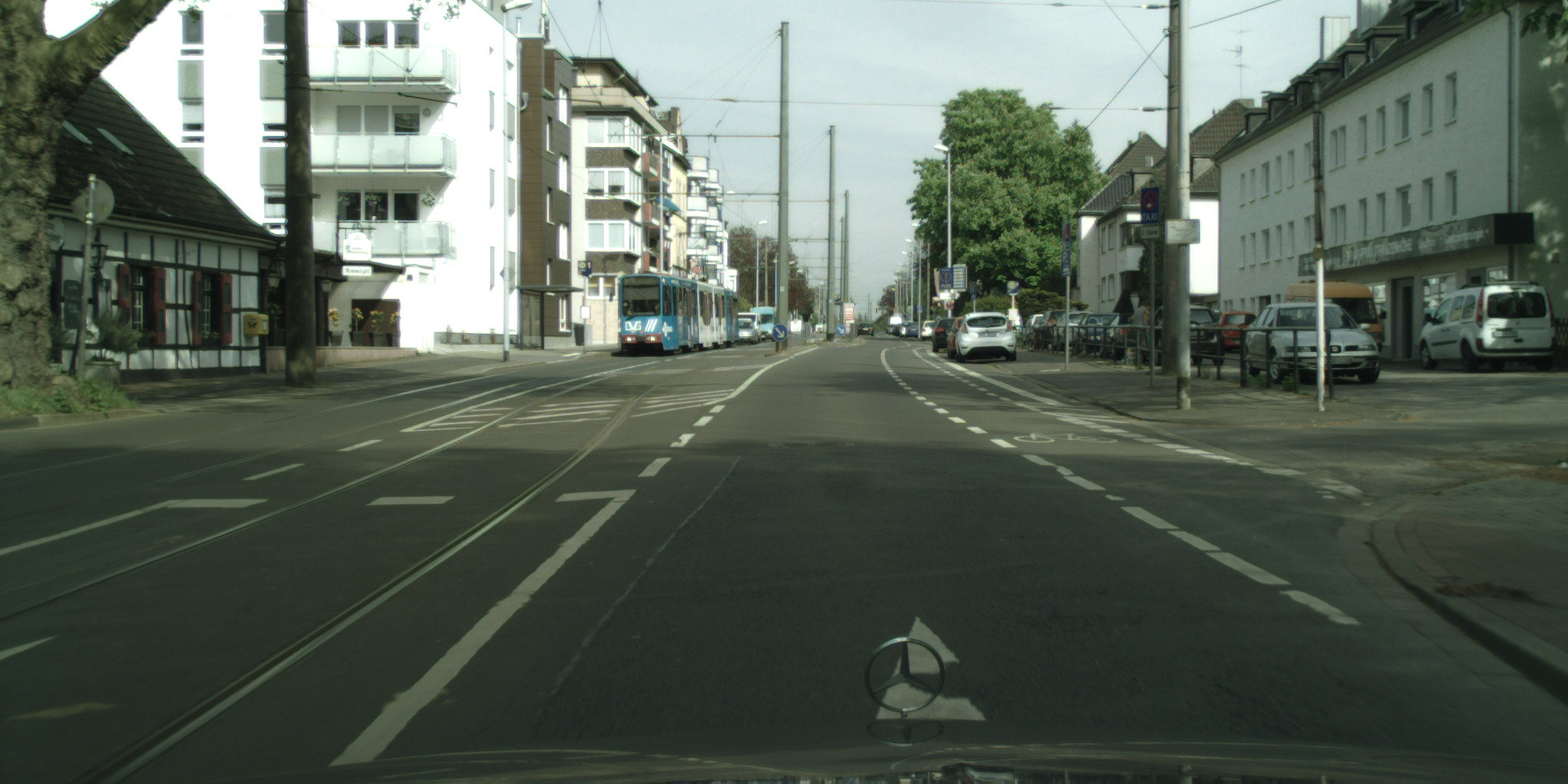}
\end{subfigure}
\hspace*{\fill}
\begin{subfigure}[b]{0.24\textwidth}
  \centering
  \includegraphics[width=\textwidth]{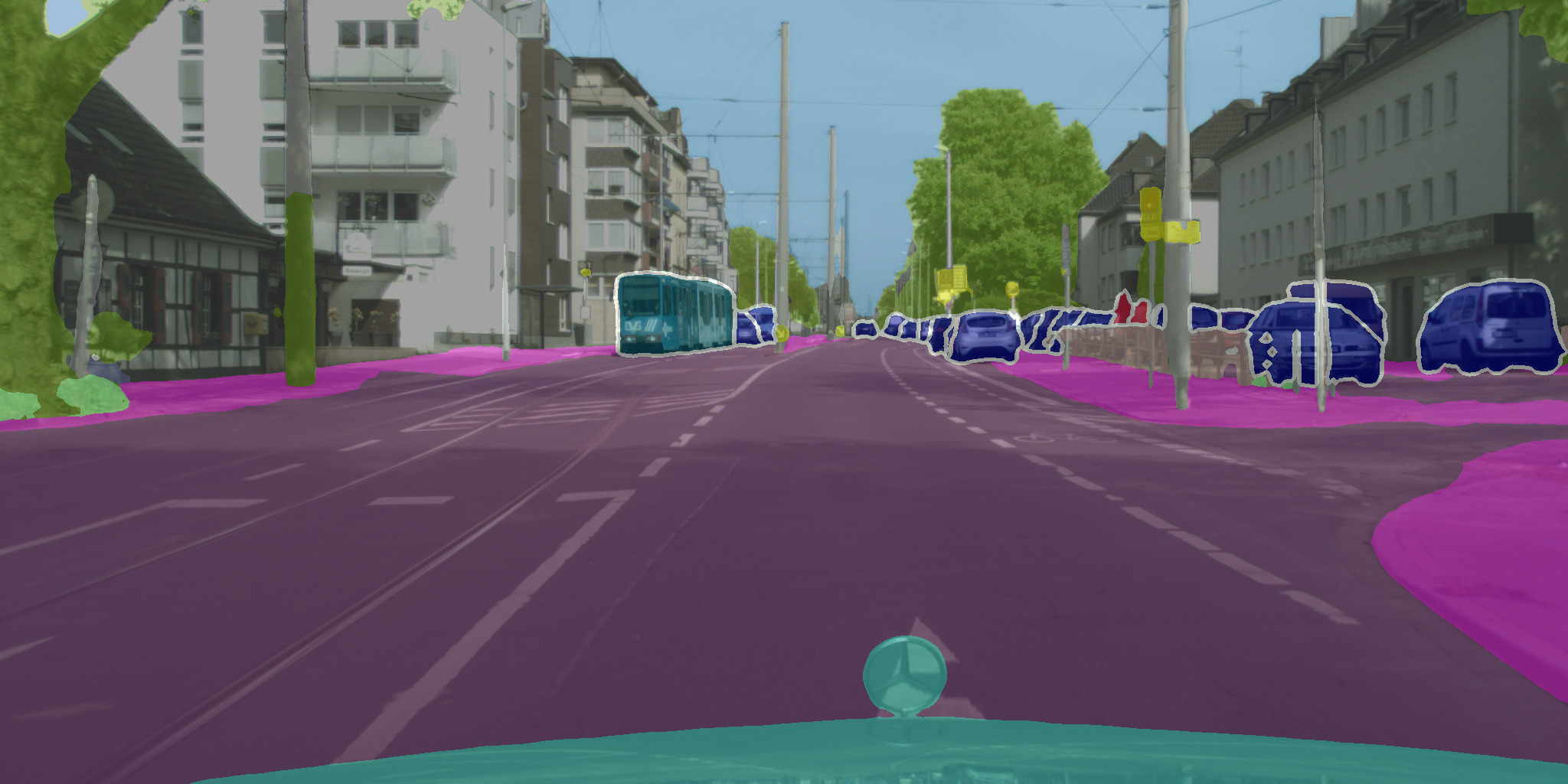}
\end{subfigure}
\begin{subfigure}[b]{0.24\textwidth}
  \centering
  \includegraphics[width=\textwidth]{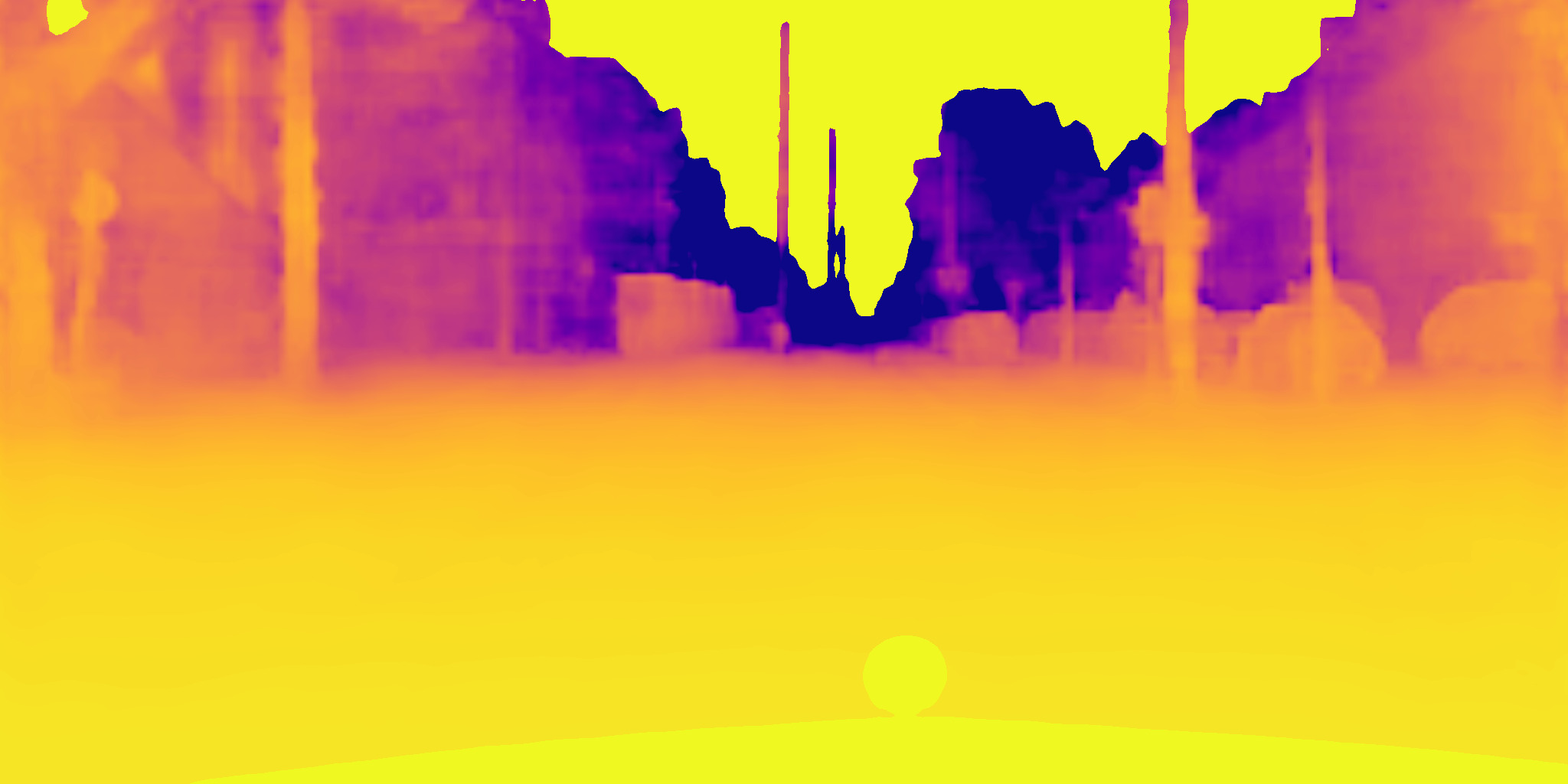}
\end{subfigure}
\hspace*{\fill}
\begin{subfigure}[b]{0.24\textwidth}
  \centering
  \includegraphics[width=\textwidth]{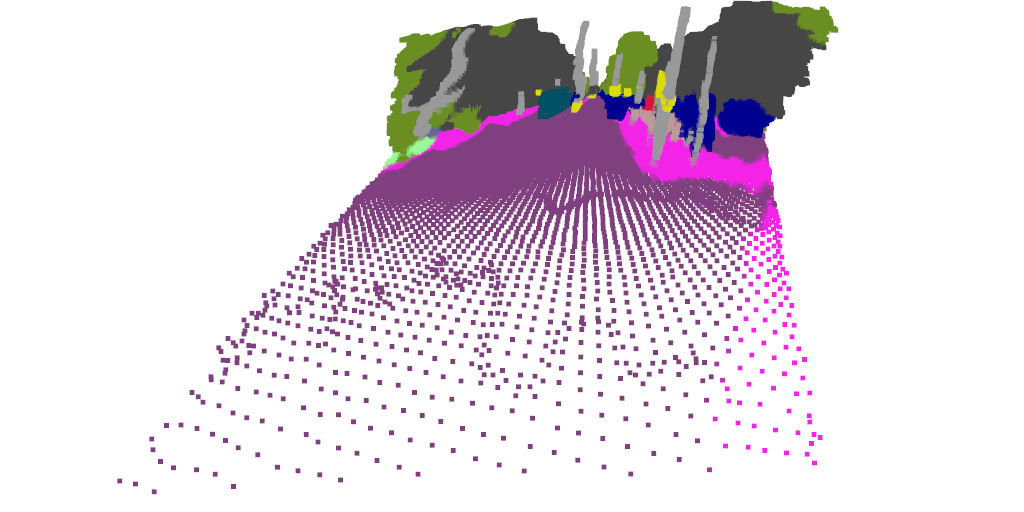}
\end{subfigure}
\vspace*{\fill}
\begin{subfigure}[b]{0.24\textwidth}
  \centering
  \includegraphics[width=\textwidth]{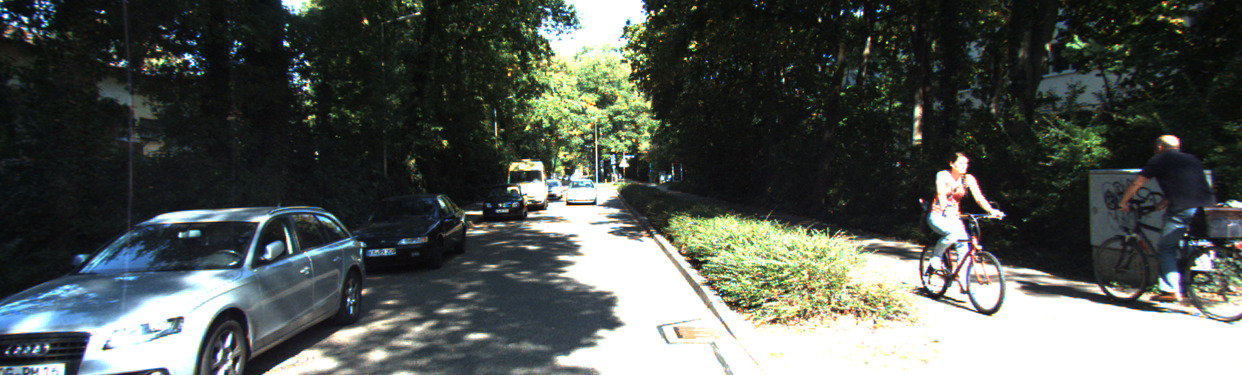}
\end{subfigure}
\hspace*{\fill}
\begin{subfigure}[b]{0.24\textwidth}
  \centering
  \includegraphics[width=\textwidth]{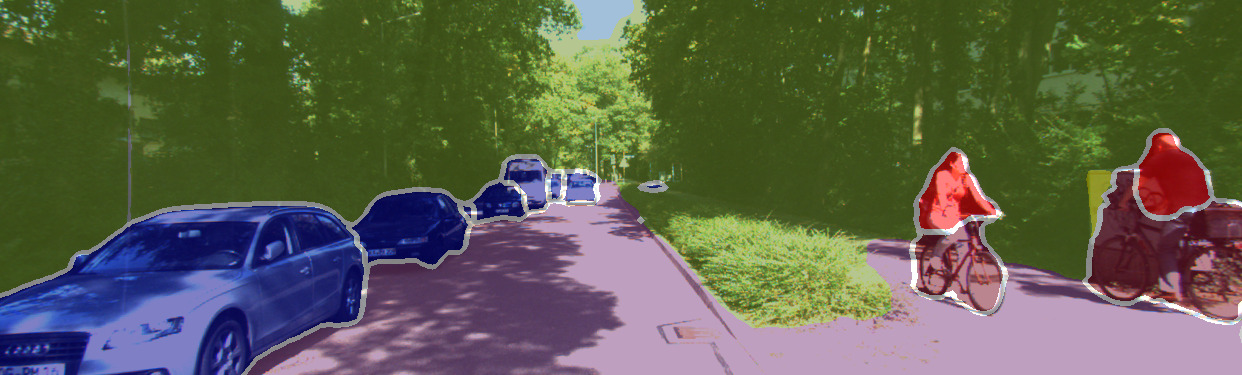}
\end{subfigure}
\begin{subfigure}[b]{0.24\textwidth}
  \centering
  \includegraphics[width=\textwidth]{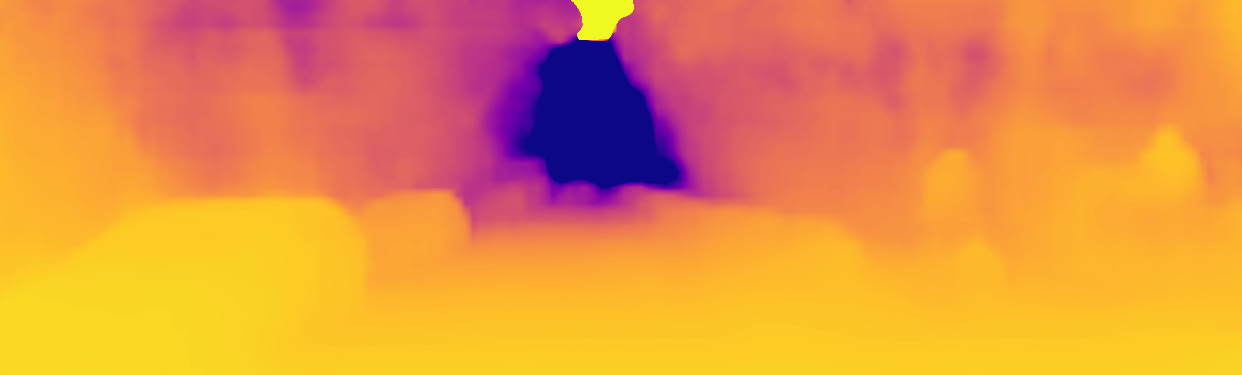}
\end{subfigure}
\hspace*{\fill}
\begin{subfigure}[b]{0.24\textwidth}
  \centering
  \includegraphics[width=\textwidth]{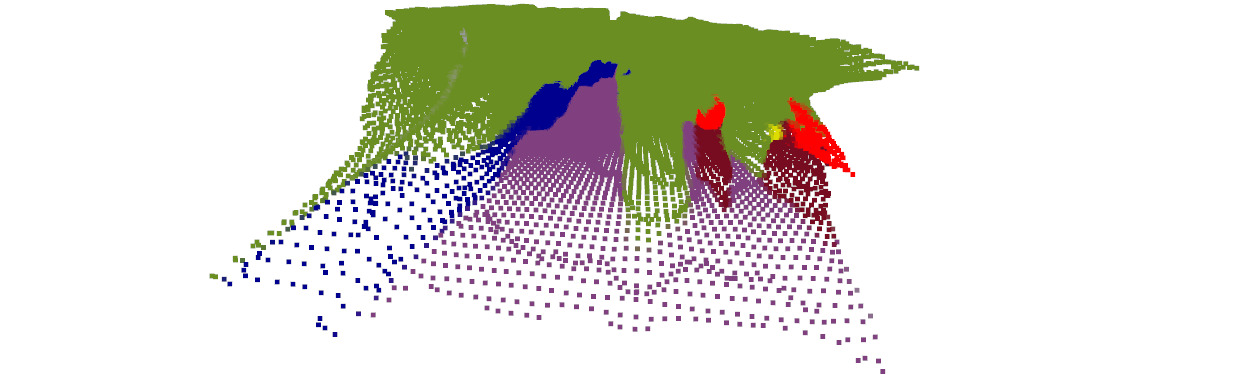}
\end{subfigure}
\vspace*{\fill}
\begin{subfigure}[b]{0.24\textwidth}
  \centering
  \includegraphics[width=\textwidth]{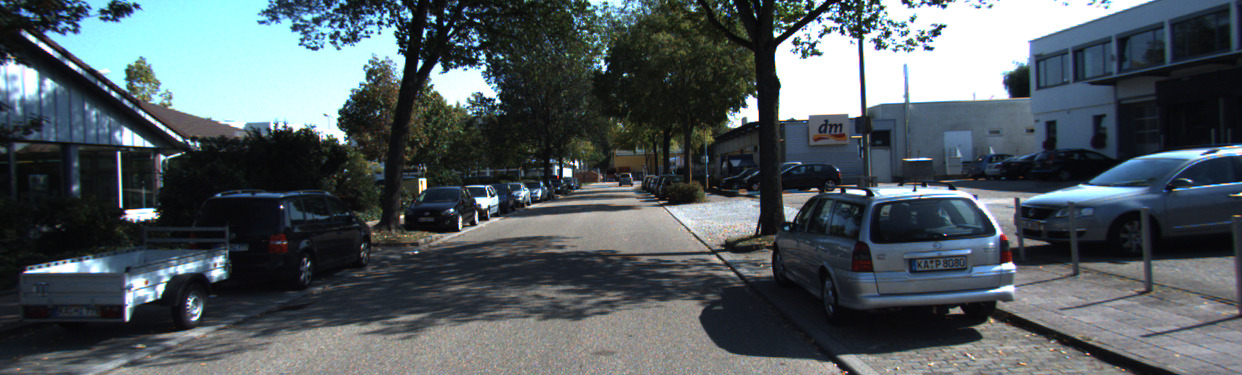}
  \caption{Input image}
\end{subfigure}
\hspace*{\fill}
\begin{subfigure}[b]{0.24\textwidth}
  \centering
  \includegraphics[width=\textwidth]{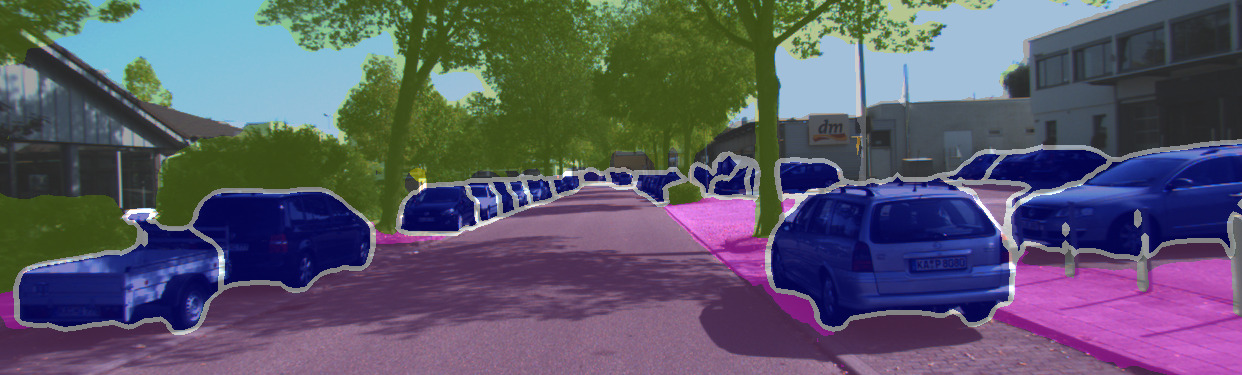}
  \caption{Panoptic segmentation}
\end{subfigure}
\begin{subfigure}[b]{0.24\textwidth}
  \centering
  \includegraphics[width=\textwidth]{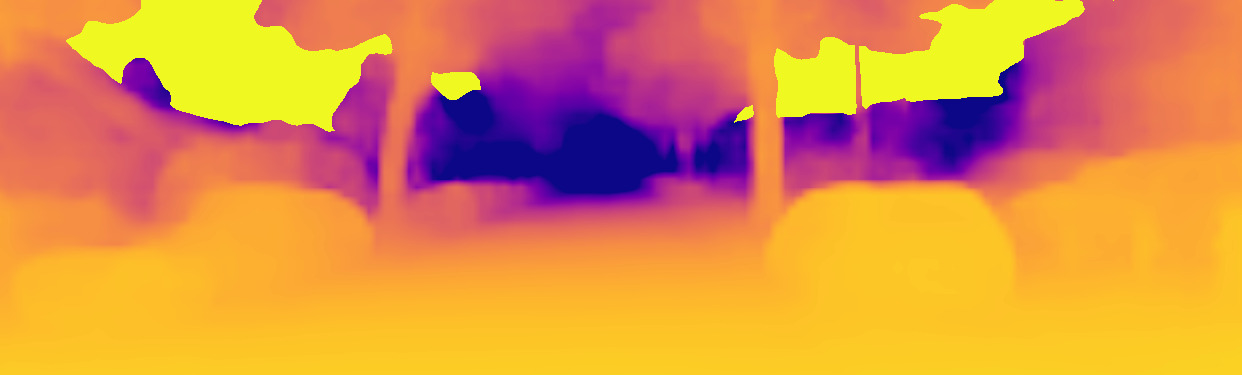}
  \caption{Monocular depth estimation}
\end{subfigure}
\hspace*{\fill}
\begin{subfigure}[b]{0.24\textwidth}
  \centering
  \includegraphics[width=\textwidth]{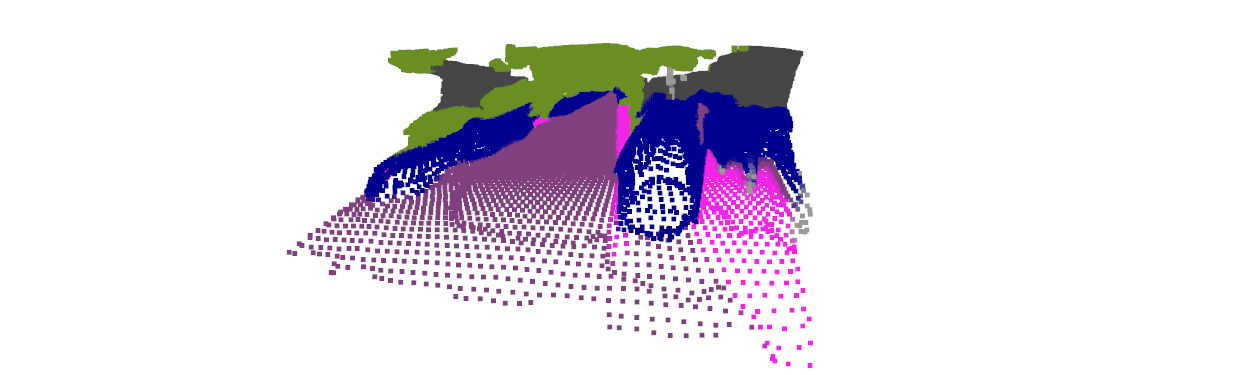}
  \caption{3D point cloud}
\end{subfigure}
\caption{Qualitative results on unseen images from the Cityscapes and KITTI dataset. The columns (from left to right) show the input image, the panoptic prediction, the monocular depth estimation, and the final 3D point cloud prediction, respectively. Instances are omitted in the 3D point clouds for better visualization. The first two rows show predictions on the Cityscapes dataset, while the last two rows show predictions on the KITTI dataset.}
\label{fig:qualitative}
\end{figure*}

\subsection{Cityscapes}
We compare our final model with state-of-the-art methods on the Cityscapes validation set in Table~\ref{tab:cityscapes} with respect to PQ and end-to-end runtime.
For a fair runtime comparison, we only infer our panoptic segmentation network part and report the unoptimized performance. 
Our model is the fastest model with a total end-to-end runtime of 44~\si{ms} and over 30~\% faster compared to the second fastest method.
The margin most likely is even higher than reported, given the fact that all other model runtimes were reported on faster GPUs.
In terms of PQ, there is a big gap to top performing methods.
This is not surprising, given the fact that these methods need more than 100~\si{ms} using much heavier architectures than our method.
Compared to Panoptic DeepLab~\cite{cheng2020panoptic} with a MobileNetV3 backbone, which is the most similar variant in literature to our method, our model provides a better speed-accuracy trade-off with 0.3~\% better PQ and over 30~\% faster runtime.
Hou~\etal~\cite{Hou_2020_CVPR} state an optimized runtime of 30~FPS in their paper, being as fast as our optimized network with a higher PQ of 58.8.
However, our model additionally predicts depth in a self-supervised fashion.
We show qualitative results of our method on unseen images from the Cityscapes dataset in Figure~\ref{fig:qualitative}.

\subsection{KITTI}
For KITTI, we use our best model from Cityscapes to generate pseudo labels on the KITTI training set.
This enables us to train on the full Eigen split and optimize both tasks in conjunction, similar to Cityscapes.
We again use additional uncertainty weighting to balance the loss terms.
Qualitative results of our method on unseen images from the KITTI dataset can be found in Figure~\ref{fig:qualitative}.
We report quantitative results for depth estimation and compare them to other self-supervised methods on the KITTI Eigen split in Table~\ref{tab:kitti}.
Compared to previous methods, our model is on par with Monodepth2~\cite{monodepth2} and exceeds most other previous methods.
Only two methods, PackNet-SfM~\cite{packnet} and SynDistNet~\cite{kumar2021syndistnet}, are able to consistently outperform our model.
PackNet-SfM uses a heavier network architecture, reaching only 6.25~FPS on full resolution $1280\times384$ pixel images.
In contrast, our method reaches 82~FPS on full resolution images and thus being real-time capable.
SynDistNet uses a multi-task approach with explicit semantic guidance.
While SynDistNet cannot perform panoptic segmentation, the model shows, that explicit methods can further boost single-task performance compared to our implicit method.
\section{Conclusion}
\label{conclusion}
In this work, we introduced the task of monocular geometric scene understanding as the combination of panoptic segmentation and self-supervised depth estimation.
Tackling this complex task, we introduced MGNet, an efficient architecture, which operates in real-time by carefully combining state-of-the-art concepts from both domains with a lightweight network architecture.
We evaluated our model on Cityscapes and KITTI and show competitive results.
On Cityscapes, we achieve 55.7~PQ and 8.3~RMSE with 30~FPS on $1024\times2048$~pixel images.
On KITTI, we achieve 3.761~RMSE with 82~FPS on $384\times1280$~pixel images.
While state-of-the-art methods provide higher accuracy on single tasks compared to our work, they are not suited for real-time applications, \eg, autonomous driving.

We hope that our work will inspire researchers to further investigate the task of monocular geometric scene understanding.
Future research can focus on using self-supervised methods for panoptic segmentation, integrating related tasks such as 3D object detection into the framework, or investigating the relation between the tasks using explicit methods to boost single-task performance.
\section*{Acknowledgment}

This research is accomplished within the project UNICARagil (FKZ 16EMO0290). We acknowledge the financial support for the project by the Federal Ministry of Education and Research of Germany (BMBF).

{\small
\bibliographystyle{ieee_fullname}
\bibliography{library}

\begin{thebibliography}{10}\itemsep=-1pt

\bibitem{caruana1997multitask}
Rich Caruana.
\newblock Multitask learning.
\newblock {\em Machine learning}, 28(1):41--75, 1997.

\bibitem{casser2019depth}
Vincent Casser, Soeren Pirk, Reza Mahjourian, and Anelia Angelova.
\newblock Depth prediction without the sensors: Leveraging structure for
  unsupervised learning from monocular videos.
\newblock In {\em Proceedings of the AAAI Conference on Artificial
  Intelligence}, volume~33, pages 8001--8008, 2019.

\bibitem{chen2020naive}
Liang-Chieh Chen, Raphael~Gontijo Lopes, Bowen Cheng, Maxwell~D Collins, Ekin~D
  Cubuk, Barret Zoph, Hartwig Adam, and Jonathon Shlens.
\newblock Naive-student: Leveraging semi-supervised learning in video sequences
  for urban scene segmentation.
\newblock In {\em Proceedings of the European conference on computer vision
  (ECCV)}, pages 695--714. Springer, 2020.

\bibitem{chen2020scaling}
Liang-Chieh Chen, Huiyu Wang, and Siyuan Qiao.
\newblock Scaling wide residual networks for panoptic segmentation.
\newblock {\em arXiv preprint arXiv:2011.11675}, 2020.

\bibitem{deeplabv3plus2018}
Liang-Chieh Chen, Yukun Zhu, George Papandreou, Florian Schroff, and Hartwig
  Adam.
\newblock Encoder-decoder with atrous separable convolution for semantic image
  segmentation.
\newblock In {\em Proceedings of the European conference on computer vision
  (ECCV)}, pages 801--818. Springer, 2018.

\bibitem{Chen_2020_CVPR}
Yifeng Chen, Guangchen Lin, Songyuan Li, Omar Bourahla, Yiming Wu, Fangfang
  Wang, Junyi Feng, Mingliang Xu, and Xi Li.
\newblock Banet: Bidirectional aggregation network with occlusion handling for
  panoptic segmentation.
\newblock In {\em Proceedings of the IEEE/CVF Conference on Computer Vision and
  Pattern Recognition (CVPR)}, 2020.

\bibitem{cheng2020panoptic}
Bowen Cheng, Maxwell~D Collins, Yukun Zhu, Ting Liu, Thomas~S Huang, Hartwig
  Adam, and Liang-Chieh Chen.
\newblock Panoptic-deeplab: A simple, strong, and fast baseline for bottom-up
  panoptic segmentation.
\newblock In {\em Proceedings of the IEEE/CVF Conference on Computer Vision and
  Pattern Recognition (CVPR)}, 2020.

\bibitem{Cordts2016Cityscapes}
Marius Cordts, Mohamed Omran, Sebastian Ramos, Timo Rehfeld, Markus Enzweiler,
  Rodrigo Benenson, Uwe Franke, Stefan Roth, and Bernt Schiele.
\newblock The cityscapes dataset for semantic urban scene understanding.
\newblock In {\em Proceedings of the IEEE Conference on Computer Vision and
  Pattern Recognition (CVPR)}, 2016.

\bibitem{DeGeus2020FastPanoptic}
D. {de Geus}, P. {Meletis}, and G. {Dubbelman}.
\newblock Fast panoptic segmentation network.
\newblock {\em IEEE Robotics and Automation Letters}, 5(2):1742--1749, 2020.

\bibitem{deng2009imagenet}
Jia Deng, Wei Dong, Richard Socher, Li-Jia Li, Kai Li, and Li Fei-Fei.
\newblock Imagenet: A large-scale hierarchical image database.
\newblock In {\em Proceedings of the IEEE Conference on Computer Vision and
  Pattern Recognition (CVPR)}, pages 248--255, 2009.

\bibitem{NIPS2014_7bccfde7}
David Eigen, Christian Puhrsch, and Rob Fergus.
\newblock Depth map prediction from a single image using a multi-scale deep
  network.
\newblock In {\em Advances in Neural Information Processing Systems (NeurIPS)},
  volume~27. Curran Associates, Inc., 2014.

\bibitem{Gao_2019_ICCV}
Naiyu Gao, Yanhu Shan, Yupei Wang, Xin Zhao, Yinan Yu, Ming Yang, and Kaiqi
  Huang.
\newblock Ssap: Single-shot instance segmentation with affinity pyramid.
\newblock In {\em Proceedings of the IEEE/CVF International Conference on
  Computer Vision (ICCV)}, 2019.

\bibitem{Gao2020LearningCA}
Naiyu Gao, Yanhu Shan, Xin Zhao, and Kaiqi Huang.
\newblock Learning category- and instance-aware pixel embedding for fast
  panoptic segmentation.
\newblock {\em IEEE Transactions on Image Processing}, 30:6013--6023, 2021.

\bibitem{garg2016unsupervised}
Ravi Garg, Vijay~Kumar Bg, Gustavo Carneiro, and Ian Reid.
\newblock Unsupervised cnn for single view depth estimation: Geometry to the
  rescue.
\newblock In {\em Proceedings of the European conference on computer vision
  (ECCV)}, pages 740--756. Springer, 2016.

\bibitem{Geiger2013IJRR}
Andreas Geiger, Philip Lenz, Christoph Stiller, and Raquel Urtasun.
\newblock Vision meets robotics: The kitti dataset.
\newblock {\em International Journal of Robotics Research (IJRR)}, 2013.

\bibitem{monodepth17}
Clement Godard, Oisin Mac~Aodha, and Gabriel~J. Brostow.
\newblock Unsupervised monocular depth estimation with left-right consistency.
\newblock In {\em Proceedings of the IEEE Conference on Computer Vision and
  Pattern Recognition (CVPR)}, 2017.

\bibitem{monodepth2}
Clement Godard, Oisin Mac~Aodha, Michael Firman, and Gabriel~J. Brostow.
\newblock Digging into self-supervised monocular depth estimation.
\newblock In {\em Proceedings of the IEEE/CVF International Conference on
  Computer Vision (ICCV)}, 2019.

\bibitem{goel2021quadronet}
Kratarth Goel, Praveen Srinivasan, Sarah Tariq, and James Philbin.
\newblock Quadronet: Multi-task learning for real-time semantic depth aware
  instance segmentation.
\newblock In {\em Proceedings of the IEEE/CVF Winter Conference on Applications
  of Computer Vision (WACV)}, pages 315--324, 2021.

\bibitem{gordon2019depth}
Ariel Gordon, Hanhan Li, Rico Jonschkowski, and Anelia Angelova.
\newblock Depth from videos in the wild: Unsupervised monocular depth learning
  from unknown cameras.
\newblock In {\em Proceedings of the IEEE/CVF International Conference on
  Computer Vision (ICCV)}, pages 8977--8986, 2019.

\bibitem{packnet}
Vitor Guizilini, Rares Ambrus, Sudeep Pillai, Allan Raventos, and Adrien
  Gaidon.
\newblock 3d packing for self-supervised monocular depth estimation.
\newblock In {\em Proceedings of the IEEE/CVF Conference on Computer Vision and
  Pattern Recognition (CVPR)}, 2020.

\bibitem{packnet-semguided}
Vitor Guizilini, Rui Hou, Jie Li, Rares Ambrus, and Adrien Gaidon.
\newblock Semantically-guided representation learning for self-supervised
  monocular depth.
\newblock In {\em International Conference on Learning Representations (ICLR)},
  2020.

\bibitem{He_2017_ICCV}
Kaiming He, Georgia Gkioxari, Piotr Dollar, and Ross Girshick.
\newblock Mask r-cnn.
\newblock In {\em Proceedings of the IEEE International Conference on Computer
  Vision (ICCV)}, 2017.

\bibitem{he2016deep}
Kaiming He, Xiangyu Zhang, Shaoqing Ren, and Jian Sun.
\newblock Deep residual learning for image recognition.
\newblock In {\em Proceedings of the IEEE conference on computer vision and
  pattern recognition (CVPR)}, pages 770--778, 2016.

\bibitem{hirschmuller2007stereo}
Heiko Hirschmuller.
\newblock Stereo processing by semiglobal matching and mutual information.
\newblock {\em IEEE Transactions on pattern analysis and machine intelligence},
  30(2):328--341, 2007.

\bibitem{homayounfar2020levelset}
Namdar Homayounfar, Yuwen Xiong, Justin Liang, Wei-Chiu Ma, and Raquel Urtasun.
\newblock Levelset r-cnn: A deep variational method for instance segmentation.
\newblock In {\em Proceedings of the European conference on computer vision
  (ECCV)}, pages 555--571. Springer, 2020.

\bibitem{Hou_2020_CVPR}
Rui Hou, Jie Li, Arjun Bhargava, Allan Raventos, Vitor Guizilini, Chao Fang,
  Jerome Lynch, and Adrien Gaidon.
\newblock Real-time panoptic segmentation from dense detections.
\newblock In {\em Proceedings of the IEEE/CVF Conference on Computer Vision and
  Pattern Recognition (CVPR)}, 2020.

\bibitem{howard2019searching}
Andrew Howard, Mark Sandler, Grace Chu, Liang-Chieh Chen, Bo Chen, Mingxing
  Tan, Weijun Wang, Yukun Zhu, Ruoming Pang, Vijay Vasudevan, et~al.
\newblock Searching for mobilenetv3.
\newblock In {\em Proceedings of the IEEE/CVF International Conference on
  Computer Vision (ICCV)}, pages 1314--1324, 2019.

\bibitem{jha2020adamt}
Ankit Jha, Awanish Kumar, Biplab Banerjee, and Subhasis Chaudhuri.
\newblock Adamt-net: An adaptive weight learning based multi-task learning
  model for scene understanding.
\newblock In {\em Proceedings of IEEE/CVF Conference on Computer Vision and
  Pattern Recognition Workshops}, pages 706--707, 2020.

\bibitem{kendall2018multi}
Alex Kendall, Yarin Gal, and Roberto Cipolla.
\newblock Multi-task learning using uncertainty to weigh losses for scene
  geometry and semantics.
\newblock In {\em Proceedings of IEEE Conference on Computer Vision and Pattern
  Recognition (CVPR)}, pages 7482--7491, 2018.

\bibitem{kingma2014adam}
Diederik~P. Kingma and Jimmy Ba.
\newblock Adam: {A} method for stochastic optimization.
\newblock In {\em International Conference on Learning Representations (ICLR)},
  2015.

\bibitem{Kirillov_2019_CVPR}
Alexander Kirillov, Ross Girshick, Kaiming He, and Piotr Dollar.
\newblock Panoptic feature pyramid networks.
\newblock In {\em Proceedings of the IEEE/CVF Conference on Computer Vision and
  Pattern Recognition (CVPR)}, 2019.

\bibitem{kirillov2019panoptic}
Alexander Kirillov, Kaiming He, Ross Girshick, Carsten Rother, and Piotr
  Doll{\'a}r.
\newblock Panoptic segmentation.
\newblock In {\em Proceedings of the IEEE/CVF Conference on Computer Vision and
  Pattern Recognition (CVPR)}, pages 9404--9413, 2019.

\bibitem{klingner2020improved}
Marvin Klingner, Andreas Bar, and Tim Fingscheidt.
\newblock Improved noise and attack robustness for semantic segmentation by
  using multi-task training with self-supervised depth estimation.
\newblock In {\em Proceedings of the IEEE/CVF Conference on Computer Vision and
  Pattern Recognition Workshops}, pages 320--321, 2020.

\bibitem{klingner2020self}
Marvin Klingner, Jan-Aike Term{\"o}hlen, Jonas Mikolajczyk, and Tim
  Fingscheidt.
\newblock Self-supervised monocular depth estimation: Solving the dynamic
  object problem by semantic guidance.
\newblock In {\em Proceedings of the European conference on computer vision
  (ECCV)}, pages 582--600. Springer, 2020.

\bibitem{kumar2021syndistnet}
Varun~Ravi Kumar, Marvin Klingner, Senthil Yogamani, Stefan Milz, Tim
  Fingscheidt, and Patrick Mader.
\newblock Syndistnet: Self-supervised monocular fisheye camera distance
  estimation synergized with semantic segmentation for autonomous driving.
\newblock In {\em Proceedings of the IEEE/CVF Winter Conference on Applications
  of Computer Vision (WACV)}, pages 61--71, 2021.

\bibitem{li2018learning}
Jie Li, Allan Raventos, Arjun Bhargava, Takaaki Tagawa, and Adrien Gaidon.
\newblock Learning to fuse things and stuff.
\newblock {\em arXiv preprint arXiv:1812.01192}, 2018.

\bibitem{li2018undeepvo}
Ruihao Li, Sen Wang, Zhiqiang Long, and Dongbing Gu.
\newblock Undeepvo: Monocular visual odometry through unsupervised deep
  learning.
\newblock In {\em IEEE international conference on robotics and automation
  (ICRA)}, pages 7286--7291, 2018.

\bibitem{li2021panopticfcn}
Yanwei Li, Hengshuang Zhao, Xiaojuan Qi, Liwei Wang, Zeming Li, Jian Sun, and
  Jiaya Jia.
\newblock Fully convolutional networks for panoptic segmentation.
\newblock In {\em Proceedings of the IEEE/CVF Conference on Computer Vision and
  Pattern Recognition (CVPR)}, pages 214--223, 2021.

\bibitem{liang2020polytransform}
Justin Liang, Namdar Homayounfar, Wei-Chiu Ma, Yuwen Xiong, Rui Hu, and Raquel
  Urtasun.
\newblock Polytransform: Deep polygon transformer for instance segmentation.
\newblock In {\em Proceedings of the IEEE/CVF Conference on Computer Vision and
  Pattern Recognition (CVPR)}, pages 9131--9140, 2020.

\bibitem{liu2019end}
Huanyu Liu, Chao Peng, Changqian Yu, Jingbo Wang, Xu Liu, Gang Yu, and Wei
  Jiang.
\newblock An end-to-end network for panoptic segmentation.
\newblock In {\em Proceedings of the IEEE/CVF Conference on Computer Vision and
  Pattern Recognition (CVPR)}, pages 6172--6181, 2019.

\bibitem{liu2015parsenet}
Wei Liu, Andrew Rabinovich, and Alexander~C Berg.
\newblock Parsenet: Looking wider to see better.
\newblock {\em arXiv preprint arXiv:1506.04579}, 2015.

\bibitem{long2015fully}
Jonathan Long, Evan Shelhamer, and Trevor Darrell.
\newblock Fully convolutional networks for semantic segmentation.
\newblock In {\em Proceedings of the IEEE Conference on Computer Vision and
  Pattern Recognition (CVPR)}, pages 3431--3440, 2015.

\bibitem{luo2019every}
Chenxu Luo, Zhenheng Yang, Peng Wang, Yang Wang, Wei Xu, Ram Nevatia, and Alan
  Yuille.
\newblock Every pixel counts++: Joint learning of geometry and motion with 3d
  holistic understanding.
\newblock {\em IEEE transactions on pattern analysis and machine intelligence},
  42(10):2624--2641, 2019.

\bibitem{maas2013rectifier}
Andrew~L Maas, Awni~Y Hannun, and Andrew~Y Ng.
\newblock Rectifier nonlinearities improve neural network acoustic models.
\newblock In {\em Proceedings of the 30th International Conference on Machine
  Learning (ICML)}, 2013.

\bibitem{mahjourian2018unsupervised}
Reza Mahjourian, Martin Wicke, and Anelia Angelova.
\newblock Unsupervised learning of depth and ego-motion from monocular video
  using 3d geometric constraints.
\newblock In {\em Proceedings of the IEEE Conference on Computer Vision and
  Pattern Recognition (CVPR)}, pages 5667--5675, 2018.

\bibitem{mohan2020efficientps}
Rohit Mohan and Abhinav Valada.
\newblock Efficientps: Efficient panoptic segmentation.
\newblock {\em International Journal of Computer Vision (IJCV)}, 2021.

\bibitem{MVD2017}
Gerhard Neuhold, Tobias Ollmann, Samuel Rota~Bul\`o, and Peter Kontschieder.
\newblock The mapillary vistas dataset for semantic understanding of street
  scenes.
\newblock In {\em Proceedings of the IEEE International Conference on Computer
  Vision (ICCV)}, 2017.

\bibitem{neven2017fast}
Davy Neven, Bert De~Brabandere, Stamatios Georgoulis, Marc Proesmans, and Luc
  Van~Gool.
\newblock Fast scene understanding for autonomous driving.
\newblock {\em arXiv preprint arXiv:1708.02550}, 2017.

\bibitem{tensorrt}
NVIDIA.
\newblock Tensorrt library.
\newblock \url{https://developer.nvidia.com/tensorrt}.

\bibitem{NEURIPS2019_9015}
Adam Paszke, Sam Gross, Francisco Massa, Adam Lerer, James Bradbury, Gregory
  Chanan, Trevor Killeen, Zeming Lin, Natalia Gimelshein, Luca Antiga, Alban
  Desmaison, Andreas Kopf, Edward Yang, Zachary DeVito, Martin Raison, Alykhan
  Tejani, Sasank Chilamkurthy, Benoit Steiner, Lu Fang, Junjie Bai, and Soumith
  Chintala.
\newblock Pytorch: An imperative style, high-performance deep learning library.
\newblock In {\em Advances in Neural Information Processing Systems (NeurIPS)},
  volume~32, pages 8024--8035. Curran Associates, Inc., 2019.

\bibitem{petrovai2020real}
Andra Petrovai and Sergiu Nedevschi.
\newblock Real-time panoptic segmentation with prototype masks for automated
  driving.
\newblock In {\em IEEE Intelligent Vehicles Symposium (IV)}, pages 1400--1406,
  2020.

\bibitem{Porzi_2021_CVPR}
Lorenzo Porzi, Samuel~Rota Bulo, and Peter Kontschieder.
\newblock Improving panoptic segmentation at all scales.
\newblock In {\em Proceedings of the IEEE/CVF Conference on Computer Vision and
  Pattern Recognition (CVPR)}, pages 7302--7311, 2021.

\bibitem{Porzi_2019_CVPR}
Lorenzo Porzi, Samuel Rota~Bul\`o, Aleksander Colovic, and Peter Kontschieder.
\newblock Seamless scene segmentation.
\newblock In {\em Proceedings of the IEEE/CVF Conference on Computer Vision and
  Pattern Recognition (CVPR)}, 2019.

\bibitem{Qiao_2021_CVPR}
Siyuan Qiao, Liang-Chieh Chen, and Alan Yuille.
\newblock Detectors: Detecting objects with recursive feature pyramid and
  switchable atrous convolution.
\newblock In {\em Proceedings of the IEEE/CVF Conference on Computer Vision and
  Pattern Recognition (CVPR)}, pages 10213--10224, 2021.

\bibitem{QiaoVip_2021_CVPR}
Siyuan Qiao, Yukun Zhu, Hartwig Adam, Alan Yuille, and Liang-Chieh Chen.
\newblock Vip-deeplab: Learning visual perception with depth-aware video
  panoptic segmentation.
\newblock In {\em Proceedings of the IEEE/CVF Conference on Computer Vision and
  Pattern Recognition (CVPR)}, pages 3997--4008, 2021.

\bibitem{rotabulo2017place}
Samuel Rota~Bul\`o, Lorenzo Porzi, and Peter Kontschieder.
\newblock In-place activated batchnorm for memory-optimized training of dnns.
\newblock In {\em Proceedings of the IEEE Conference on Computer Vision and
  Pattern Recognition (CVPR)}, 2018.

\bibitem{saeedan2021boosting}
Faraz Saeedan and Stefan Roth.
\newblock Boosting monocular depth with panoptic segmentation maps.
\newblock In {\em Proceedings of the IEEE/CVF Winter Conference on Applications
  of Computer Vision (WACV)}, pages 3853--3862, 2021.

\bibitem{Sofiiuk_2019_ICCV}
Konstantin Sofiiuk, Olga Barinova, and Anton Konushin.
\newblock Adaptis: Adaptive instance selection network.
\newblock In {\em Proceedings of the IEEE/CVF International Conference on
  Computer Vision (ICCV)}, 2019.

\bibitem{tan2019mnasnet}
Mingxing Tan, Bo Chen, Ruoming Pang, Vijay Vasudevan, Mark Sandler, Andrew
  Howard, and Quoc~V Le.
\newblock Mnasnet: Platform-aware neural architecture search for mobile.
\newblock In {\em Proceedings of the IEEE/CVF Conference on Computer Vision and
  Pattern Recognition (CVPR)}, pages 2820--2828, 2019.

\bibitem{tan2019efficientnet}
Mingxing Tan and Quoc Le.
\newblock Efficientnet: Rethinking model scaling for convolutional neural
  networks.
\newblock In {\em International Conference on Machine Learning}, pages
  6105--6114. PMLR, 2019.

\bibitem{Tao2020HierarchicalMA}
A. Tao, K. Sapra, and Bryan Catanzaro.
\newblock Hierarchical multi-scale attention for semantic segmentation.
\newblock {\em arXiv preprint arXiv:2005.10821}, 2020.

\bibitem{Uhrig2017THREEDV}
Jonas Uhrig, Nick Schneider, Lukas Schneider, Uwe Franke, Thomas Brox, and
  Andreas Geiger.
\newblock Sparsity invariant cnns.
\newblock In {\em International Conference on 3D Vision (3DV)}, 2017.

\bibitem{vandenhende2020mti}
Simon Vandenhende, Stamatios Georgoulis, and Luc Van~Gool.
\newblock Mti-net: Multi-scale task interaction networks for multi-task
  learning.
\newblock In {\em Proceedings of the European conference on computer vision
  (ECCV)}, pages 527--543. Springer, 2020.

\bibitem{wang2018learning}
Chaoyang Wang, Jos{\'e}~Miguel Buenaposada, Rui Zhu, and Simon Lucey.
\newblock Learning depth from monocular videos using direct methods.
\newblock In {\em Proceedings of the IEEE Conference on Computer Vision and
  Pattern Recognition (CVPR)}, pages 2022--2030, 2018.

\bibitem{wang2020max}
Huiyu Wang, Yukun Zhu, Hartwig Adam, Alan Yuille, and Liang-Chieh Chen.
\newblock Max-deeplab: End-to-end panoptic segmentation with mask transformers.
\newblock In {\em Proceedings of the IEEE/CVF Conference on Computer Vision and
  Pattern Recognition (CVPR)}, pages 5463--5474, 2021.

\bibitem{wang2020axial}
Huiyu Wang, Yukun Zhu, Bradley Green, Hartwig Adam, Alan Yuille, and
  Liang-Chieh Chen.
\newblock Axial-deeplab: Stand-alone axial-attention for panoptic segmentation.
\newblock In {\em Proceedings of the European conference on computer vision
  (ECCV)}, pages 108--126. Springer, 2020.

\bibitem{WangSCJDZLMTWLX19}
Jingdong Wang, Ke Sun, Tianheng Cheng, Borui Jiang, Chaorui Deng, Yang Zhao,
  Dong Liu, Yadong Mu, Mingkui Tan, Xinggang Wang, Wenyu Liu, and Bin Xiao.
\newblock Deep high-resolution representation learning for visual recognition.
\newblock {\em IEEE transactions on pattern analysis and machine intelligence},
  2020.

\bibitem{wang2020sdc}
Lijun Wang, Jianming Zhang, Oliver Wang, Zhe Lin, and Huchuan Lu.
\newblock Sdc-depth: Semantic divide-and-conquer network for monocular depth
  estimation.
\newblock In {\em Proceedings of IEEE/CVF Conference on Computer Vision and
  Pattern Recognition (CVPR)}, pages 541--550, 2020.

\bibitem{wang2020solo}
Xinlong Wang, Tao Kong, Chunhua Shen, Yuning Jiang, and Lei Li.
\newblock Solo: Segmenting objects by locations.
\newblock In {\em Proceedings of the European conference on computer vision
  (ECCV)}, pages 649--665. Springer, 2020.

\bibitem{wang2020solov2}
Xinlong Wang, Rufeng Zhang, Tao Kong, Lei Li, and Chunhua Shen.
\newblock Solov2: Dynamic and fast instance segmentation.
\newblock In {\em Advances in Neural Information Processing Systems (NeurIPS)},
  volume~33, pages 17721--17732. Curran Associates, Inc., 2020.

\bibitem{wang2004image}
Zhou Wang, Alan~C Bovik, Hamid~R Sheikh, and Eero~P Simoncelli.
\newblock Image quality assessment: from error visibility to structural
  similarity.
\newblock {\em IEEE transactions on image processing}, 13(4):600--612, 2004.

\bibitem{xiong2019upsnet}
Yuwen Xiong, Renjie Liao, Hengshuang Zhao, Rui Hu, Min Bai, Ersin Yumer, and
  Raquel Urtasun.
\newblock Upsnet: A unified panoptic segmentation network.
\newblock In {\em Proceedings of the IEEE/CVF Conference on Computer Vision and
  Pattern Recognition (CVPR)}, pages 8818--8826, 2019.

\bibitem{Xiong_2019_CVPR}
Yuwen Xiong, Renjie Liao, Hengshuang Zhao, Rui Hu, Min Bai, Ersin Yumer, and
  Raquel Urtasun.
\newblock Upsnet: A unified panoptic segmentation network.
\newblock In {\em Proceedings of the IEEE/CVF Conference on Computer Vision and
  Pattern Recognition (CVPR)}, 2019.

\bibitem{xu2018pad}
Dan Xu, Wanli Ouyang, Xiaogang Wang, and Nicu Sebe.
\newblock Pad-net: Multi-tasks guided prediction-and-distillation network for
  simultaneous depth estimation and scene parsing.
\newblock In {\em Proceedings of IEEE Conference on Computer Vision and Pattern
  Recognition (CVPR)}, pages 675--684, 2018.

\bibitem{xue2020toward}
F. {Xue}, G. {Zhuo}, Z. {Huang}, W. {Fu}, Z. {Wu}, and M.~H. {Ang}.
\newblock Toward hierarchical self-supervised monocular absolute depth
  estimation for autonomous driving applications.
\newblock In {\em IEEE/RSJ International Conference on Intelligent Robots and
  Systems (IROS)}, pages 2330--2337, 2020.

\bibitem{yang2019deeperlab}
Tien-Ju Yang, Maxwell~D Collins, Yukun Zhu, Jyh-Jing Hwang, Ting Liu, Xiao
  Zhang, Vivienne Sze, George Papandreou, and Liang-Chieh Chen.
\newblock Deeperlab: Single-shot image parser.
\newblock {\em arXiv preprint arXiv:1902.05093}, 2019.

\bibitem{yin2018geonet}
Zhichao Yin and Jianping Shi.
\newblock Geonet: Unsupervised learning of dense depth, optical flow and camera
  pose.
\newblock In {\em Proceedings of the IEEE conference on computer vision and
  pattern recognition (CVPR)}, pages 1983--1992, 2018.

\bibitem{yu2018bisenet}
Changqian Yu, Jingbo Wang, Chao Peng, Changxin Gao, Gang Yu, and Nong Sang.
\newblock Bisenet: Bilateral segmentation network for real-time semantic
  segmentation.
\newblock In {\em Proceedings of the European conference on computer vision
  (ECCV)}, pages 325--341. Springer, 2018.

\bibitem{zamir2020robust}
Amir~R Zamir, Alexander Sax, Nikhil Cheerla, Rohan Suri, Zhangjie Cao, Jitendra
  Malik, and Leonidas~J Guibas.
\newblock Robust learning through cross-task consistency.
\newblock In {\em Proceedings of IEEE/CVF Conference on Computer Vision and
  Pattern Recognition (CVPR)}, pages 11197--11206, 2020.

\bibitem{zhang2019pattern}
Zhenyu Zhang, Zhen Cui, Chunyan Xu, Yan Yan, Nicu Sebe, and Jian Yang.
\newblock Pattern-affinitive propagation across depth, surface normal and
  semantic segmentation.
\newblock In {\em Proceedings of IEEE/CVF Conference on Computer Vision and
  Pattern Recognition (CVPR)}, pages 4106--4115, 2019.

\bibitem{zhao2017pyramid}
Hengshuang Zhao, Jianping Shi, Xiaojuan Qi, Xiaogang Wang, and Jiaya Jia.
\newblock Pyramid scene parsing network.
\newblock In {\em Proceedings of the IEEE Conference on Computer Vision and
  Pattern Recognition (CVPR)}, pages 2881--2890, 2017.

\bibitem{zhou2020constant}
Hang Zhou, David Greenwood, Sarah Taylor, and Han Gong.
\newblock Constant velocity constraints for self-supervised monocular depth
  estimation.
\newblock In {\em European Conference on Visual Media Production}, pages 1--8,
  2020.

\bibitem{zhou2019unsupervised}
Junsheng Zhou, Yuwang Wang, Kaihuai Qin, and Wenjun Zeng.
\newblock Unsupervised high-resolution depth learning from videos with dual
  networks.
\newblock In {\em Proceedings of the IEEE/CVF International Conference on
  Computer Vision (ICCV)}, pages 6872--6881, 2019.

\bibitem{zhou2017unsupervised}
Tinghui Zhou, Matthew Brown, Noah Snavely, and David~G Lowe.
\newblock Unsupervised learning of depth and ego-motion from video.
\newblock In {\em Proceedings of the IEEE Conference on Computer Vision and
  Pattern Recognition (CVPR)}, pages 1851--1858, 2017.

\bibitem{semantic_cvpr19}
Yi Zhu, Karan Sapra, Fitsum~A. Reda, Kevin~J. Shih, Shawn Newsam, Andrew Tao,
  and Bryan Catanzaro.
\newblock Improving semantic segmentation via video propagation and label
  relaxation.
\newblock In {\em Proceedings of the IEEE/CVF Conference on Computer Vision and
  Pattern Recognition (CVPR)}, 2019.

\bibitem{zou2018df}
Yuliang Zou, Zelun Luo, and Jia-Bin Huang.
\newblock Df-net: Unsupervised joint learning of depth and flow using
  cross-task consistency.
\newblock In {\em Proceedings of the European conference on computer vision
  (ECCV)}, pages 36--53. Springer, 2018.

\end{thebibliography}
}
\clearpage

\includepdf[pages=1]{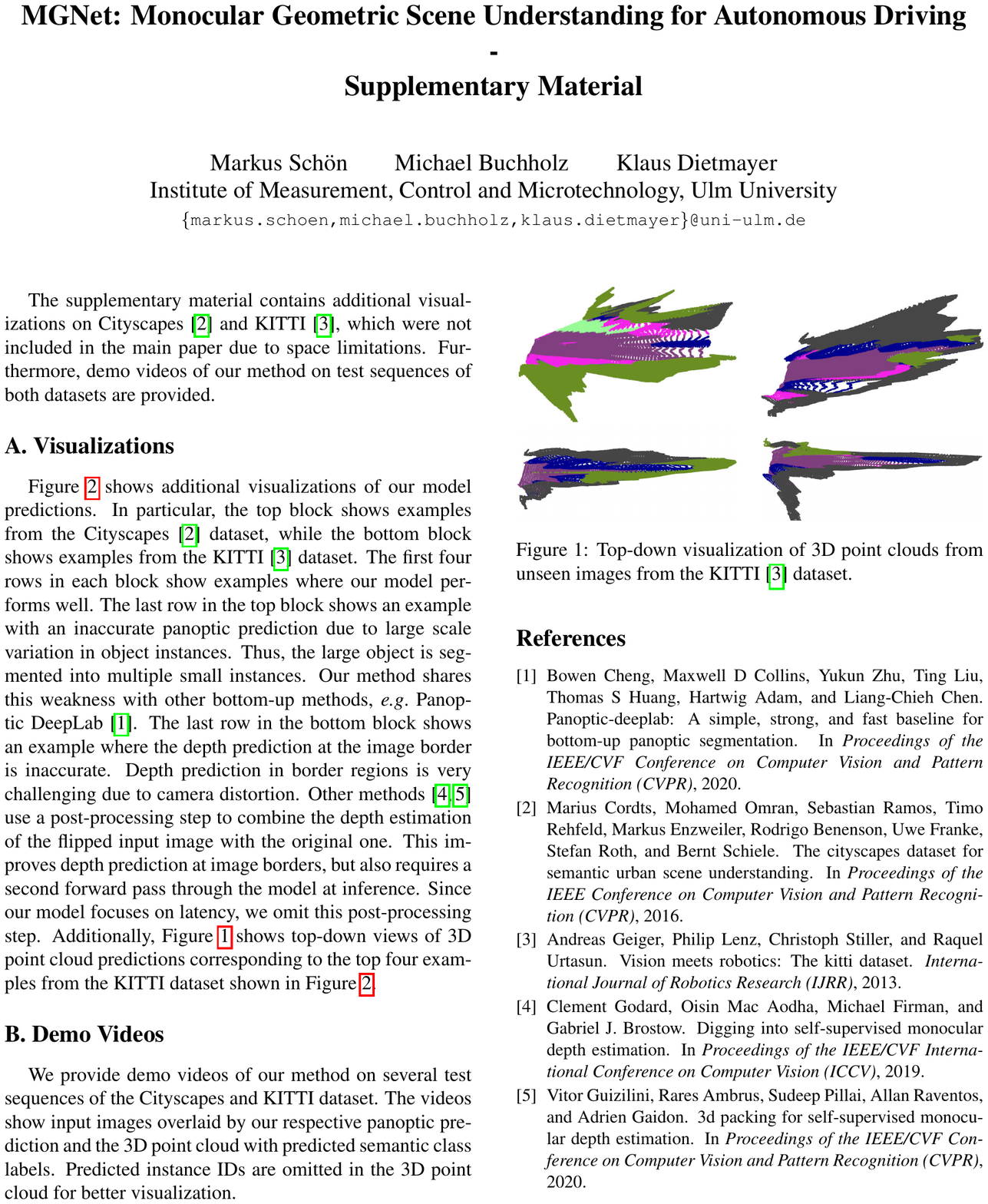}
\includepdf[pages=2]{img/supplement.pdf}

\end{document}